\def\year{2022}\relax
\documentclass[letterpaper]{article} 
\usepackage{aaai22}  
\usepackage{times}  
\usepackage{helvet}  
\usepackage{courier}  
\usepackage[hyphens]{url}  
\usepackage{graphicx} 
\usepackage{natbib}  
\usepackage{caption} 
\DeclareCaptionStyle{ruled}{labelfont=normalfont,labelsep=colon,strut=off} 
\usepackage{multicol}

\usepackage[utf8]{inputenc}

\usepackage{mathrsfs} 
\usepackage{amssymb,amsmath,amsthm,graphicx,float}
\usepackage{mathtools}
\usepackage{booktabs}
\usepackage{tabularx}
\newcommand*{\arial}{\fontfamily{phv}\selectfont}
\usepackage{caption}
\usepackage{subcaption}

\usepackage{algorithm}
\usepackage{algpseudocode}

\newtheorem{problem}{Problem}

\newtheorem{remark}{Remark}

\DeclareMathOperator*{\argmax}{arg\,max}

\newcommand{\Pomdp}{\mathcal{M}}

\newcommand{\States}{\mathcal{S}}
\newcommand{\state}{s}
\newcommand{\action}{\alpha}
\newcommand{\Actions}{\mathcal{A}}
\newcommand{\Transition}{\mathcal{P}}
\newcommand{\Observations}{\mathcal{O}}
\newcommand{\ObservationSet}{\mathcal{Z}}
\newcommand{\observation}{z}

\newcommand{\StateOccup}{\mu}
\newcommand{\StateActionOccup}{\nu}

\newcommand{\Reward}{\mathcal{R}}
\newcommand{\Initdist}{\StateOccup_0}
\newcommand{\discount}{\gamma}
\newcommand{\Real}{\mathbb{R}}

\newcommand{\Distr}{\mathrm{Distr}}
\newcommand{\policy}{\sigma}
\newcommand{\Rewardvec}{\theta}
\newcommand{\trustregion}{\rho}
\newcommand{\slack}{k}
\newcommand{\mapping}{\eta}
\newcommand{\memoryupt}{\delta}
\newcommand{\target}{\mathcal{T}}

\newcommand{\reachPropSymbol}{\varphi}
\usepackage{tikz,pgfplots}
\pgfplotsset{compat=1.8}
\usepackage{pgfplotstable}
\usepgfplotslibrary{groupplots}
\usepackage[framemethod=default]{mdframed}

\usepackage{times}
\usepackage{bbm}
\usepgfplotslibrary{fillbetween}
\usetikzlibrary{arrows,decorations.pathmorphing,positioning,fit,trees,shapes,shadows,automata,calc} 
\usetikzlibrary{patterns,arrows,arrows.meta,calc,shapes,shadows,decorations.pathmorphing,decorations.pathreplacing,automata,shapes.multipart,positioning,shapes.geometric,fit,circuits,trees,shapes.gates.logic.US,fit}
\usetikzlibrary{backgrounds,scopes}
\usepackage{aligned-overset}

 \newenvironment{customlegend}[1][]{%
        \begingroup
        \csname pgfplots@init@cleared@structures\endcsname
        \pgfplotsset{#1}%
    }{%
        \csname pgfplots@createlegend\endcsname
        \endgroup
    }%

    \def\addlegendimage{\csname pgfplots@addlegendimage\endcsname}
 \newcommand\titlesize{\fontsize{8.1pt}{10.2pt}\selectfont}

\usepackage{newfloat}
\usepackage{listings}
\floatstyle{ruled}
\newfloat{listing}{tb}{lst}{}
\floatname{listing}{Listing}

\usepackage{xcolor} 

\title{Task-Guided Inverse Reinforcement Learning Under Partial Information}
\author{
    Franck Djeumou\textsuperscript{\rm 1},
    Murat Cubuktepe\textsuperscript{\rm 1},
    Craig Lennon\textsuperscript{\rm 2},
    and Ufuk Topcu\textsuperscript{\rm 1}
}
\affiliations{
    \textsuperscript{\rm 1} The University of Texas at Austin \\
    \textsuperscript{\rm 2} Army Research Laboratory \\
    fdjeumou@utexas.edu, mcubuktepe@utexas.edu, craig.t.lennon.civ@mail.mil, utopcu@utexas.edu

}

\begin{document}


\maketitle

\begin{abstract}
We study the problem of inverse reinforcement learning (IRL), where the learning agent recovers a reward function using expert demonstrations.
Most of the existing IRL techniques make the often unrealistic assumption that the agent has access to full information about the environment.
We remove this assumption by developing an algorithm for IRL in partially observable Markov decision processes (POMDPs).
The algorithm addresses several limitations of existing techniques that do not take the \emph{information asymmetry} between the expert and the learner into account.
First, it adopts causal entropy as the measure of the likelihood of the expert demonstrations as opposed to entropy in most existing IRL techniques, and avoids a common source of algorithmic complexity.
Second, it incorporates task specifications expressed in temporal logic into IRL.
Such specifications may be interpreted as side information available to the learner a priori in addition to the demonstrations and may reduce the information asymmetry.
Nevertheless, the resulting formulation is still nonconvex due to the intrinsic nonconvexity of the so-called \emph{forward problem}, i.e., computing an optimal policy given a reward function, in POMDPs.
We address this nonconvexity through sequential convex programming and introduce several extensions to solve the forward problem in a scalable manner.
This scalability allows computing policies that incorporate memory at the expense of added computational cost yet also outperform memoryless policies.
We demonstrate that, even with severely limited data, the algorithm learns reward functions and policies that satisfy the task and induce a similar behavior to the expert by leveraging the side information and incorporating memory into the policy.
\end{abstract}
\section{Introduction}
Inverse reinforcement learning (IRL) is a technique that recovers a reward function using expert demonstrations and learns a policy inducing a similar behavior to the expert's.
IRL techniques have found a wide range of applications~\cite{abbeel2010autonomous,kitani2012activity,hadfield2016cooperative,dragan2013policy,finn2016guided}.
The majority of the work has focused on Markov decision processes (MDPs), assuming that the learning agent can fully observe the state of the environment and the expert's demonstrations~\cite{abbeel2010autonomous,ziebart2008maximum,zhou2017infinite,ziebart2010modeling,hadfield2016cooperative,finn2016guided}.
Often, in reality, the learning agent will not have such full observation.
For example, a robot will never know everything about its environment~\cite{ong2009pomdps,bai2014integrated,zhang2017robot} and may not observe the internal states of a human with whom it works~\cite{akash2019improving,liu2012modeling}.
Such information limitations violate the intrinsic assumptions made in existing IRL techniques.

We study IRL in partially observable Markov decision processes (POMDPs), a widely used model for decision-making under imperfect information. 
The partial observability brings three key challenges in IRL.
The first two challenges are related to the so-called \emph{information asymmetry} between the expert and the learner. 
First, the expert typically has access to full information about the environment, while the learner has only a partial view of the expert's demonstrations.
Second, even in the hypothetical case in which the actual reward function is known to the learner, the learner's optimal policy under limited information may not yield the same behavior as the expert due to information asymmetry.

The third challenge is due to the computational complexity of policy synthesis in POMDPs.
Many standard IRL techniques rely on a subroutine that solves the so-called \emph{forward problem}, i.e., computing an optimal policy for a given reward.
Solving the forward problem for POMDPs is significantly more challenging than MDPs, both theoretically and practically. 
Optimal policies for POMDPs may require infinite memory of observations~\cite{MadaniHC99}, whereas memoryless policies are enough for MDPs.

An additional limitation in existing IRL techniques is due to the limited expressivity and often impracticability of state-based reward functions in representing complex tasks~\cite{littman2017environment}.
For example, it will be tremendously difficult to define a merely state-based reward function to describe requirements such as ``do not steer off the road while reaching the target location and coming back to home'' or ``monitor multiple locations with a certain order.''
On the other hand, such requirements can be concisely and precisely specified in temporal logic~\cite{BK08,Pnueli}.
Recent work has demonstrated the utility of incorporating temporal logic specifications into IRL in MDPs~\cite{memarian2020active,wen2017learning}. In this work, we address these challenges and limitations in IRL techniques by studying the problem:
\begin{mdframed}[backgroundcolor=gray!30,  linecolor=black!60]
\textbf{Task-guided IRL:} Given a POMDP, a \emph{task specification}  $\varphi$ expressed in temporal logic, and a set of expert demonstrations, learn a policy along with the underlying reward function that maximizes the \emph{causal entropy} of the induced stochastic process, induces a behavior similar to the expert's, and ensures satisfaction of $\varphi$.
\end{mdframed}%

We highlight two parts of the problem statement.
Using \emph{causal entropy} as an optimization criterion results in a least-committal policy that induces a behavior obtaining the same accumulated reward as the expert's demonstrations while making no additional assumptions about the demonstrations.
Given the task requirements, the \emph{task specifications} guide the learning process by describing the feasible behaviors and allowing to learn performant policies with respect to the task requirements.
Such specifications can also be interpreted as side information available to the learner a priori in addition to the demonstrations and partially alleviates the information asymmetry between the expert and the learner.

Most existing work on IRL relies on \emph{entropy} as a measure of the likelihood of the demonstrations, yet,  when applied to stochastic MDPs, has to deal with nonconvex optimization problems~\cite{ziebart2008maximum,ziebart2010modeling}.
On the other hand, IRL techniques that adopt \emph{causal entropy} as the measure of likelihood enjoy formulations based on convex optimization~\cite{zhou2017infinite,ziebart2010modeling}.
We show similar algorithmic benefits in maximum-causal-entropy IRL carry over from MDPs to POMDPs.

A major difference between MDPs and POMDPs in maximum-causal-entropy IRL is, though, due to the intrinsic nonconvexity of policy synthesis in POMDPs, which yields a formulation of the task-guided IRL problem as a nonconvex optimization.
It is known that this nonconvexity severely limits the scalability for synthesis in POMDPs.
We develop an algorithm that solves the resulting nonconvex problem in a scalable manner by adapting sequential convex programming (SCP)~\cite{yuan2015recent,mao2018successive}.
The algorithm is iterative.
In each iteration, it linearizes the underlying nonconvex problem around the solution from the previous iteration. The algorithm introduces several extensions, among which a verification step not present in existing SCP schemes.
We show that it computes a sound and locally optimal solution to the task-guided IRL problem.

In several examples, we show that the algorithm scales to POMDPs with tens of thousands of states as opposed to tens of states in the existing work, e.g., belief-based techniques.
In POMDPs, \emph{finite-memory} policies that are functions of the history of the observations outperform memoryless policies~\cite{yu2008near}. Computing a finite-memory policy for a POMDP is equivalent to computing a memoryless policy on a larger product POMDP~\cite{junges2018finite}.
On the other hand, existing IRL techniques on POMDPs cannot effectively utilize memory, as they do not scale to large POMDPs.
We leverage the scalability of our algorithm to compute more performant policies that incorporate memory using finite-state controllers~\cite{meuleau1999solving,amato2010optimizing}.

We demonstrate the applicability of the approach through several examples. We show that, without task specifications, the developed algorithm can compute more performant policies than a straight adaptation of the original GAIL~\cite{ho2016generative} to POMDPs. Then, we demonstrate that by incorporating task specifications into the IRL procedure, the learned reward function and policy accurately describe the behavior of the expert while outperforming the policy obtained without the task specifications.
Additionally, we show that incorporating memory into the learning agent's policy leads to more performant policies.
We also show that with more limited data, the performance gap becomes more prominent between the learned policies with and without using task specifications.
Finally, we demonstrate the scalability of our approach for solving the \emph{forward problem} through extensive comparisons with several state-of-the-art POMDP solvers and show that on larger POMDPs, the algorithm can compute more performant policies in significantly less time.

\paragraph{Related work.} 
The closest work to ours is by~\citet{choi2011inverse}, where they extend classical maximum-margin-based IRL techniques for MDPs to POMDPs. However, even on MDPs, maximum-margin-based approaches cannot resolve the ambiguity caused by suboptimal demonstrations, and they work well when there is a single reward function that is clearly better than alternatives~\cite{osa2018algorithmic}. In contrast, we adopt causal entropy that has been shown~\cite{osa2018algorithmic, ziebart2010modeling} to alleviate these limitations on MDPs. Besides, ~\citet{choi2011inverse} rely on efficient off-the-shelf solvers to the forward problem. Instead, this paper also develops an algorithm that outperforms off-the-shelf solvers and can scale to POMDPs that are orders of magnitude larger compared to the examples in ~\citet{choi2011inverse}. Further, \citet{choi2011inverse} do not incorporate task specifications in their formulations.

Prior work tackled the ill-posed IRL problem using maximum margin formulations~\cite{ratliff2006maximum,abbeel2004apprenticeship,ng2000algorithms}, or probabilistic models to compute the likelihood of expert demonstrations~\cite{ramachandran2007bayesian,ziebart2008maximum,ziebart2010modeling,zhou2017infinite, finn2016guided, ho2016generative}. Besides, the idea of using side information to guide and augment IRL has been explored in recent work~\cite{papusha2018inverse,wen2017learning,memarian2020active}.
However, these IRL techniques are only applicable to MDPs as opposed to POMDPs.

IRL under some restricted notion of partial information has been studied in prior work. \citet{boularias2012structured} consider the setting where the features of the reward function are partially specified. \citet{kitani2012activity,bogert2014multi} consider IRL problems from partially observable demonstrations and use the hidden Markov decision process framework as a solution. Therefore, all these approaches consider a particular case of POMDPs. We also note that none of these methods incorporate side information into IRL and do not provide guarantees on the performance of the policy with respect to a task specification.
\section{Background}
\paragraph{Notation.} We denote the set of nonnegative real numbers by $\Real_+$, the set of all probability distributions over a finite or countably infinite set $\mathcal{X}$ by $\Distr(\mathcal{X})$, the set of all (infinite or empty) sequences $x_0,x_1,\hdots,x_\infty$ with $x_i \in \mathcal{X}$ by $(\mathcal{X})^*$ for some set $\mathcal{X}$, and the expectation of a function $g$ of jointly distributed random variables $X$ and $Y$ by $\mathbb{E}_{X,Y}[g(X,Y)]$.%

\paragraph{POMDPs.} A partially observable Markov decision process (POMDP) is a tuple $\Pomdp = (\States, \Actions,  \ObservationSet, \Transition, \Observations,\Reward, \Initdist, \discount)$, with finite sets $\States$, $\Actions$ and $\ObservationSet$ denoting the set of states, actions, and observations, respectively, a transition function $\Transition : \States \times \Actions \mapsto \Distr(\States)$, an observation function $\Observations : \States \mapsto \Distr(\ObservationSet)$, a reward function $\Reward: \States \times \Actions \mapsto \Real_+$, an initial state of distribution $\Initdist \in \Distr(\States)$, and a discount factor $\discount \in (0, 1)$.
We denote $\Transition(\state' | \state,\action)$ as the probability of transitioning to state $\state'$ after an action $\action$ is selected in state $\state$, and $\Observations(\observation | \state)$ is the probability of observing $\observation\in\ObservationSet$ in state $\state$. 

\paragraph{Policies.}
An observation-based policy $\policy : (\ObservationSet \times \Actions)^* \times \ObservationSet \mapsto \Distr(\Actions)$ for a POMDP $\Pomdp$ maps a sequence of observations and actions to a distribution over actions.
A $\mathrm{M}$-\emph{finite-state controller} ($\mathrm{M}$-FSC) consists of a finite set of memory states of size $\mathrm{M}$ and two functions. 
The \emph{action mapping} $\mapping(n,\observation)$ takes a FSC memory state $n$ and an observation $\observation \in \ObservationSet$, and returns a distribution over the POMDP actions.
The \emph{memory update} $\memoryupt(n,\observation,\action)$ returns a distribution over memory states and is a function of the action $\action$ selected by $\mapping$. An FSC induces an observation-based policy by following a joint execution of these two functions upon a trace of the POMDP. \emph{Memoryless FSCs, denoted by $\policy\colon \ObservationSet \rightarrow\Distr(\Actions)$, are observation-based policies, where $\policy_{\observation,\action}$ is the probability of taking the action $\action$ given solely observation $\observation$.}

\begin{remark}[\textsc{Reduction to Memoryless Policies}]\label{rem:mempolicy}
In the remainder of the paper, for ease of notation, we synthesize optimal $\mathrm{M}$-FSCs for POMDPs (so-called forward problem) by computing memoryless policies $\policy$ on theoretically-justified larger POMDPs obtained from the so-called product of the memory update $\memoryupt$ and the original POMDPs. Indeed,~\citet{junges2018finite} provide product POMDPs, whose sizes grow polynomially with the size of the domain of $\memoryupt$.
\end{remark}

\paragraph{Causal Entropy in POMDPs.}
For a POMDP $\Pomdp$, a policy $\policy$ induces the stochastic processes $S^\policy_{0:\infty} := (S^\policy_0,\hdots,S^\policy_\infty) $, $A^\policy_{0:\infty} := (A^\policy_0,\hdots,A^\policy_\infty)$, and $Z^\policy_{0:\infty} := (Z^\policy_0,\hdots,Z^\policy_\infty)$. 
At each time index $t$, the random variables $S^\policy_t$, $A^\policy_t$, and $Z^\policy_t$ take values $\state_t \in \States$, $\action_t \in \Actions$, and $\observation_t \in \ObservationSet$, respectively. 

\emph{We consider the infinite time horizon setting and define the discounted causal entropy as~\cite{zhou2017infinite}: $H_\policy^\discount  :=  \sum\nolimits_{t=0}^\infty \discount^t \mathbb{E}_{A^\policy_t,S^\policy_t}[- \log \mathbb{P}(A^\policy_t | S^\policy_t)]$.}
\begin{remark}
The \emph{entropy} of POMDPs (or MDPs) involves the \emph{future policy decisions}~\cite{ziebart2008maximum}, i.e., $S^\policy_{t+1:T}$, at a time index $t$, as opposed to the causal entropy in POMDPs (or MDPs). Thus, \citet{ziebart2008maximum} show that the problem of computing a policy that maximizes the entropy is nonconvex, even in MDPs.
Inverse reinforcement learning techniques that maximize the entropy of the policy rely on approximations or assume that the transition function of the MDP is deterministic.
On the other hand, computing a policy that maximizes the causal entropy can be formulated as a convex optimization problem in MDPs~\cite{ziebart2010modeling,zhou2017infinite}.
\end{remark}

\paragraph{LTL Specifications.} We use general linear temporal logic (LTL) to express complex task specifications on the POMDP $\Pomdp$. Given a set $\mathrm{AP}$ of atomic propositions, i.e., Boolean variables with truth values for a given state $s$ or observation $z$, LTL formulae are constructed inductively as following: $\reachPropSymbol := \mathrm{true} \;|\; a \;|\; \lnot \reachPropSymbol \;|\; \reachPropSymbol_1 \wedge \reachPropSymbol_2 \;|\; \textbf{X} \reachPropSymbol \;|\; \reachPropSymbol_1 \textbf{U} \reachPropSymbol_2$, where $a \in \mathrm{AP}$, $\reachPropSymbol$, $\reachPropSymbol_1$, and $\reachPropSymbol_2$ are LTL formulae, $\lnot$ and $\wedge$ are the logic negation and conjunction, and $\textbf{X}$ and $\textbf{U}$ are the \emph{next} and \emph{until} temporal operators. Besides, temporal operators such as \emph{always} ($\textbf{G}$) and \emph{eventually} ($\textbf{F}$) are derived as $\textbf{F}\reachPropSymbol := \mathrm{true} \textbf{U} \reachPropSymbol$ and $\textbf{G} \reachPropSymbol := \lnot \textbf{F} \lnot\reachPropSymbol$. A detailed description of the syntax and semantics of LTL is beyond the scope of this paper and can be found in~\citet{Pnueli,BK08}. 

\emph{$\mathrm{Pr}_{\Pomdp}^\policy(\varphi)$ denotes the probability of satisfying the LTL formula $\varphi$ when following the policy $\sigma$ on the POMDP $\Pomdp$.}
\section{Formal Problem Statement}
In this section, we formulate the problem of task-guided inverse reinforcement learning (IRL) in POMDPs. Given a POMDP $\Pomdp$ with an \emph{unknown} reward function $\Reward$, we seek to learn a reward function $\Reward$ along with an underlying policy $\policy$ that induces a behavior similar to the expert demonstrations. 

We define an expert trajectory on the POMDP $\Pomdp$ as the perceived observation and executed action sequence $\tau = \{(\observation_0,\action_0),(\observation_1,\action_1),\hdots,(\observation_{T},\action_{T} )\}$, where $\observation_i \in \ObservationSet$ and $\action_i \in \Actions$ for all $i \in \{0,\hdots,T\}$, and $T$ denotes the length of the trajectory. Similarly to \citet{choi2011inverse}, we assume given or we can construct from $\tau$ (Bayesian inference), the belief trajectory $b^{\tau} = \{b_{0}:=\mu_0,\hdots,b_{T}\}$, where $b_{i}(\state)$ is the probability of being at state $\state$ at time index $i$. In the following, we assume that we are given a set of belief trajectories $\mathcal{D} = \{b^{\tau_1},\hdots, b^{\tau_N}\}$ from trajectories $\tau_1,\hdots,\tau_N$, where $N$ denotes the total number of underlying trajectories.

We build on the traditional encoding of the reward function as  $\Reward(\state,\action) := \sum_{k=1}^d \Rewardvec_k \phi_k(\state,\action) = \Rewardvec^\mathrm{T} \phi(\state,\action),$
where $\phi : \States \times \Actions \mapsto \Real^d$ is a known vector of basis functions with components referred to as \emph{feature functions}, $\Rewardvec \in \Real^d$ is an unknown weight vector characterizing the importance of each feature, and $d$ is the number of features.

Specifically, we seek for a weight $\Rewardvec$ defining $\Reward$ and a policy $\policy$ such that its discounted feature expectation $R_\policy^\phi$ matches an empirical discounted feature expectation $\bar{R}^\phi$ of the expert demonstration $\mathcal{D}$. That is, we have that $R_\policy^\phi = \bar{R}^\phi$, where $R_\policy^\phi := \sum_{t=0}^\infty \discount^t \mathbb{E}_{S^\policy_t,A^\policy_t}[\phi(S^\policy_t,A^\policy_t)| \policy]$ and the empirical mean $\bar{R}^\phi = \frac{1}{N} \sum_{b^{\tau} \in \mathcal{D}} \sum_{b_i \in b^{\tau} } \discount^i \sum_{\state \in \States}b_i(s) \phi(s,\action_i)$.

However, there may be infinitely many reward functions and policies that can satisfy this feature matching condition.
Therefore, to resolve the policy ambiguity, we seek for a policy $\policy$ that also maximizes the discounted causal entropy $H_\policy^\discount$. 
We now define the problem of interest.%
\begin{problem}\label{pb:irl-pomdp}
    Given a reward-free POMDP $\Pomdp$, a demonstration set $\mathcal{D}$, and a feature $\phi$, compute a policy $\policy$ and weight $\Rewardvec$ such that (a) $\sum_{t=0}^\infty \discount^t \mathbb{E}_{S^\policy_t,A^\policy_t}[\phi(S^\policy_t,A^\policy_t)| \policy] = \bar{R}^\phi$; (b) The causal entropy $H_\policy^\discount $ is maximized by $\policy$.
\end{problem}%
Additionally, we seek to incorporate, if available, a priori high-level side information on the task demonstrated by the expert in the design of the reward $\Reward$ and the policy $\policy$.%
\begin{problem}\label{pbirl-pomdp-side-info}
    Given a linear temporal logic formula $\varphi$, compute a policy $\policy$ and weight $\Rewardvec$ such that the constraints (a) and (b) in Problem~\ref{pb:irl-pomdp} are satisfied, and $\mathrm{Pr}_{\Pomdp}^\policy(\varphi) \geq \lambda$ for a given parameter $\lambda \geq 0$.
\end{problem}

We note that, even though the parameter $\lambda$ that specifies the threshold for satisfaction of $\reachPropSymbol$ is assumed to be given, the approach can easily be adapted to compute the optimal $\lambda$.
\section{Nonconvex Formulation for IRL in POMDPs}
In this section, we formulate the IRL problem and the constraints induced by (a) and (b) in Problem~\ref{pb:irl-pomdp} and Problem ~\ref{pbirl-pomdp-side-info} as a nonconvex optimization problem.
Then, we utilize a \emph{Lagrangian relaxation} of the nonconvex problem as a part of our solution approach. We recall the reader (see Remark~\ref{rem:mempolicy}) that we compute $\mathrm{M}$-FSC for POMDPs by computing memoryless policies $\policy$ on larger product POMDPs.

\paragraph{Substituting Visitation Counts.} We eliminate the (infinite) time dependency in $H_\policy^{\discount}$ and the feature matching constraint by a substitution of variables involving the policy-induced \emph{discounted state visitation count} $\StateOccup^\discount_{\policy} : \States \mapsto \Real_+$ and \emph{state-action visitation count} $\StateActionOccup^\discount_{\policy} : \States \times \Actions \mapsto \Real_+$. 
For a policy $\policy$, state $\state$, and action $\action$, the discounted visitation counts are defined by $\StateOccup^\discount_\policy (\state) := \mathbb{E}_{S_t} [\sum_{t=1}^\infty \discount^t \mathbbm{1}_{\{S_t = \state\}} | \policy]$ and $\StateActionOccup^\discount_\policy (\state,\action) := \mathbb{E}_{A_t, S_t} [\sum_{t=1}^\infty \discount^t \mathbbm{1}_{\{S_t = \state, A_t = \action\}} | \policy],$ where $\mathbbm{1}_{\{\cdot\}}$ is the indicator function. Further, $\StateActionOccup^\discount_\policy(\state,\action) = \pi_{\state,\action}\StateOccup^\discount_\policy(\state)$, where $\pi_{\state,\action} = \mathbb{P}[A_t=a | S_t = s] $ is a state-based policy.

We first provide a concave expression for the discounted causal entropy $H_\policy^\discount$ as a function of $\StateOccup^\discount_\policy$ and $\StateActionOccup^\discount_\policy$:%
\begin{align}
    H_\policy^\discount 
    &:= \sum\nolimits_{t=0}^\infty \discount^t \mathbb{E}_{S^\policy_t,A^\policy_t}[-\log(\pi_{\state_t,\action_t})] \nonumber \\
    &= \sum\nolimits_{t=0}^\infty \sum\nolimits_{(\state,\action) \in \States \times \Actions} -(\log \pi_{\state,\action}) \pi_{\state,\action} \discount^t \mathbb{P}[S^\policy_t=s] \nonumber\\
&= \sum\nolimits_{(\state,\action)\in \States \times \Actions}  -(\log \pi_{\state,\action}) \pi_{\state,\action} \StateOccup^\discount_\policy (\state) \nonumber\\
&= \sum\nolimits_{(\state,\action)\in \States \times \Actions} -\log \frac{\StateActionOccup^\discount_\policy(\state,\action)}{\StateOccup^\discount_\policy(\state)} \StateActionOccup^\discount_\policy(\state,\action), \label{eq:causal-entropy}
\end{align}%
where the first equality is due to the definition of the discounted causal entropy $H_\policy^\discount$, the second equality obtained by expanding the expectation. 
The third and fourth equalities follow by the definition of the state visitation count $\StateOccup^\discount_{\policy}$, and the state-action visitation count $\StateActionOccup^\discount_{\policy}$.
Next, we obtain a \emph{linear} expression in $\StateActionOccup^\discount_\policy$ for the discounted feature expectation $R^\policy_\phi$ as:%
\begin{align}
    R_\policy^\phi \: &= \: \sum_{t=0}^\infty \sum_{(\state,\action) \in \States \times \Actions} \phi(\state,\action) \discount^t \mathbb{P}[S^\policy_t = \state, A^\policy_t = \action] \nonumber \\
    &= \: \sum_{(\state,\action) \in \States \times \Actions} \phi(\state,\action) \StateActionOccup^\discount_\policy(\state,\action), \label{eq:feature-matching}
\end{align} where the second equality is obtained by the definition of the visitation count $\StateActionOccup^\discount_{\policy}$. The following \emph{nonconvex} constraint in $\StateOccup^\discount_\policy(\state)$ and $\policy_{\observation,\action}$ ensures observation-based policies:%
\begin{align}
 \StateActionOccup^\discount_\policy (\state,\action)  =  \StateOccup^\discount_\policy (\state) \sum\nolimits_{\observation \in \ObservationSet}\Observations(\observation|\state)\policy_{\observation,\action}. \label{eq:policy-constraint}
\end{align}%
Finally, the variables for the discounted visitation counts must satisfy the so-called \emph{Bellman flow constraint} to ensure that the policy is well-defined. For each state $s \in \States$,%
\begin{align}
    \StateOccup^\discount_{\policy} (\state)  = \Initdist(\state) + \discount \sum_{\state' \in \States} \sum_{\action \in \Actions} \Transition(\state | \state',\action) \StateActionOccup^\discount_{\policy} (\state',\action). \label{eq:bellman-constraint}
\end{align}%

\algdef{SE}[DOWHILE]{Do}{doWhile}{\algorithmicdo}[1]{\algorithmicwhile\ #1}%
\begin{algorithm}[t]
    \caption{Compute the weight vector $\Rewardvec$ and policy $\policy$ solution of the Lagrangian relaxation of the IRL problem. }\label{algo:gradweight}
    \begin{algorithmic}[1]    
        \Require{Feature expectation $\bar{R}^\phi$ from $\mathcal{D}$, initial weight $\Rewardvec^0$, step size $\eta : \mathbb{N} \mapsto \mathbb{R}^+$, and (if available) a priori side information $\varphi$ and $\lambda \in [0,1]$ imposing $\mathrm{Pr}_{\Pomdp}^\policy(\varphi) \geq \lambda$ .}
        \State $\policy^0 \gets$ uniform policy \Comment{Initialize uniform policy}
        \For{$k = 1,2,\hdots, $} \Comment{Compute $\Rewardvec$ via gradient descent}
            \State $\sigma^k \gets \texttt{SCPForward}(\Rewardvec^{k},\policy^{k-1}, \varphi, \lambda)$ \Comment{Solve the \emph{forward problem}~\eqref{eq:cost-entropy-npb}--\eqref{eq:pos-mu-state-npb} with optional $\varphi$ and $\lambda$ }
            \State $\Rewardvec^{k+1} \gets \Rewardvec^{k} - \eta(k) \big(R^\phi_{\policy^k} - \bar{R}^\phi \big)$ \Comment{Gradient step}
        \EndFor
    \State \Return $\policy^k$, $\Rewardvec^{k}$
  \end{algorithmic}
\end{algorithm}

\paragraph{Lagrangian Relaxation of Feature Matching Constraint.} Computing a policy $\policy$ that satisfies the feature matching constraint $R_\policy^\phi = \bar{R}^\phi$ might be infeasible due to $\bar{R}^\phi$ being an empirical estimate from the finite set of demonstrations $\mathcal{D}$. 
Additionally, the feature matching constraint might also be infeasible due to the information asymmetry between the expert and the learner, e.g., the expert has full observation.

We build on a \emph{Lagrangian relaxation} to incorporate the feature matching constraints into the objective of the forward problem, similar as other IRL algorithms in the literature. 
Specifically, we introduce $\Rewardvec \in \mathbb{R}^d$ as the dual variables of the relaxed problem. 
The desired weight vector $\Rewardvec$ and policy $\policy$ of Problem~\ref{pb:irl-pomdp} and Problem~\ref{pbirl-pomdp-side-info} are the solutions of $\mathrm{min}_\Rewardvec \: f(\Rewardvec) := \mathrm{max}_\policy \: H_\policy^\discount + \Rewardvec^\mathrm{T} (R_\policy^\phi - \bar{R}^\phi)$. 
Algorithm~\ref{algo:gradweight} updates the reward weights by using gradient descent.
To this end, the algorithm computes the gradient $\nabla f (\Rewardvec^{k}) = R^\phi_{\policy^k} - \bar{R}^\phi$, where $\policy^k = \argmax_\policy H_\policy^\discount + (\Rewardvec^k)^\mathrm{T} (R_\policy^\phi - \bar{R}^\phi)$.
In the following, we refer to the problem of computing such $\policy^k$ given $\Rewardvec^k$ as the \emph{forward problem}, and we develop the algorithm $\texttt{SCPForward}$, presented in next section, to solve it in an efficient and scalable manner while incorporating high-level task specifications to guide the learning.

\paragraph{Nonconvex Formulation of the Forward Problem.} Given a weight vector $\Rewardvec^k$, we take advantage of the obtained substitution by the expected visitation counts to formulate the \emph{forward problem} associated to Problem~\ref{pb:irl-pomdp} as the nonconvex optimization problem:
\begin{align}
    &\displaystyle \underset{\StateOccup^\discount_\policy,\StateActionOccup^\discount_\policy,\policy}{\mathrm{maximize}}   \: \sum_{(\state,\action)\in \States \times \Actions} -\log \frac{\StateActionOccup^\discount_\policy(\state,\action)}{\StateOccup^\discount_\policy(\state)} \StateActionOccup^\discount_\policy(\state,\action) \nonumber\\ 
    &\quad \quad \quad \quad \quad + \sum_{(\state,\action) \in \States \times \Actions} (\Rewardvec^k)^\mathrm{T} \phi(\state,\action) \StateActionOccup^\discount_\policy(\state,\action), \label{eq:cost-entropy-npb} \\
    &\displaystyle \mathrm{subject \ to} \quad\eqref{eq:policy-constraint}-\eqref{eq:bellman-constraint},\nonumber\\
    &\displaystyle\forall (\state,\action) \in \States \times \Actions, \;\; \StateOccup^\discount_{\policy} (\state) \geq 0,  \;\; \StateActionOccup^\discount_\policy (\state,\action) \geq 0, \label{eq:pos-and-mu-state-npb}\\
    & \displaystyle\forall (\state,\action) \in \States \times \Actions, \;\; \StateOccup^\discount_{\policy}(\state) = \sum\nolimits_{\action \in \Actions} \StateActionOccup^\discount_{\policy}(\state,\action), \label{eq:pos-mu-state-npb}
\end{align}%
where the source of nonconvexity is from~\eqref{eq:policy-constraint}, and we remove the constant $-(\Rewardvec^k)^\mathrm{T} \bar{R}^\phi$ from the cost function.
\section{Sequential Convex Programming Formulation}
We develop $\texttt{SCPForward}$, adapting a sequential convex programming (SCP) scheme to efficiently solve the nonconvex \emph{forward problem}~\eqref{eq:cost-entropy-npb}--\eqref{eq:pos-mu-state-npb}.
$\texttt{SCPForward}$ involves a \emph{verification step} to compute sound policies and visitation counts, which is not present in the existing SCP schemes. Additionally, we describe in the next section how to take advantage of high-level task specification (Problem~\ref{pbirl-pomdp-side-info}) through slight modifications of the obtained optimization problem solved by $\texttt{SCPForward}$.

\subsection{Linearizing Nonconvex Problem}
$\texttt{SCPForward}$ iteratively linearizes the nonconvex constraints in~\eqref{eq:policy-constraint} around a previous solution.
However, the linearization may result in an infeasible or unbounded linear subproblem~\cite{mao2018successive}. 
We first add \emph{slack variables} to the linearized constraints to ensure feasibility.
The linearized problem may not accurately approximate the nonconvex problem if the solutions to this problem deviate significantly from the previous solution.
Thus, we utilize trust region constraints~\cite{mao2018successive} to ensure that the linearization is accurate to the nonconvex problem. 
At each iteration, we introduce a \emph{verification step} to ensure that the computed policy and visitation counts are not just approximations but actually satisfy the nonconvex policy constraint~\eqref{eq:policy-constraint}, improves the realized cost function over past iterations, and satisfy the temporal logic specifications, if available.

\paragraph{Linearizing Nonconvex Constraints and Adding Slack Variables.} We linearize the nonconvex constraint~\eqref{eq:policy-constraint}, which is quadratic in $\StateOccup^\discount_\policy(\state)$ and $\policy_{\observation,\action}$, around the previously computed solution denoted by $\hat{\policy}$, $\StateOccup^\discount_{\hat{\policy}}$, and $\StateActionOccup^\discount_{\hat{\policy}}$. 
However, the linearized constraints may be infeasible.
We alleviate this drawback by adding \emph{slack variables} $\slack_{\state,\action}\in \Real$ for $(\state,\action) \in \States \times \Actions$, which results in the affine constraint:%
\begin{align}
 \StateActionOccup^\discount_\policy (\state,\action) + \slack_{\state,\action} =& \;\StateOccup^\discount_{\hat{\policy}} (\state) \sum\nolimits_{\observation \in \ObservationSet}\Observations(\observation|\state){\policy}_{\observation,\action} \;+ \label{eq:convex-policy-constraint-slack} \\
    &\big(\StateOccup^\discount_\policy (\state)-\StateOccup^\discount_{\hat{\policy}} (\state)\big) \sum\nolimits_{\observation \in \ObservationSet}\Observations(\observation|\state)\hat{\policy}_{\observation,\action}. \nonumber
\end{align}%

\paragraph{Trust Region Constraints.} The linearization may be inaccurate if the solution deviates significantly from the previous solution.
We add following \emph{trust region} constraints to alleviate this drawback:%
\begin{align}
     \forall (\observation,\action) \in \ObservationSet \times \Actions,& \quad \hat{\policy}_{\observation,\action}/\trustregion \leq {\policy}_{\observation,\action} \leq  \hat{\policy}_{\observation,\action} \trustregion, \label{eq:convex-policy-constraint-trust-region}
\end{align}%
where $\trustregion$ is the size of the trust region to restrict the set of allowed policies in the linearized problem. We augment the cost function in \eqref{eq:cost-entropy-npb} with the term $-\beta \sum_{(\state,\action) \in \States\times\Actions}\slack_{\state,\action}$ to ensure that we minimize the violation of the linearized constraints, where $\beta$ is a large positive constant.

\paragraph{Linearized Problem.} Finally, by differentiating $x \mapsto x \log x$ and $y \mapsto x \log (x/y)$, we obtain the coefficients required to linearize the convex causal entropy cost function in~\eqref{eq:cost-entropy-npb}. Thus, we obtain the following linear program (LP):%
\begin{align}
    & \underset{\StateOccup^\discount_\policy,\StateActionOccup^\discount_\policy,\policy}{\mathrm{maximize}}   \: \sum\nolimits_{(\state,\action)\in \States \times \Actions} -\Bigg(\beta\slack_{\state,\action}- \Big(\frac{\StateActionOccup^\discount_{\hat{\policy}}(\state,\action)}{\StateOccup^\discount_{\hat{\policy}}(\state)} \Big)\StateOccup^\discount_\policy(\state) \nonumber \\ 
    & \quad \quad \quad \quad \quad \: + \Big(\log \frac{\StateActionOccup^\discount_{\hat{\policy}}(\state,\action)}{\StateOccup^\discount_{\hat{\policy}}(\state)}+1 \Big)\StateActionOccup^\discount_\policy(\state,\action) \Bigg) \nonumber\\
    & \qquad \qquad \quad \: +  \sum\nolimits_{(\state,\action) \in \States \times \Actions} (\Rewardvec^k)^\mathrm{T} \phi(\state,\action) \StateActionOccup^\discount_\policy(\state,\action) \label{eq:cost-linearized-pb}\\
    & \mathrm{subject \ to} \quad  \eqref{eq:bellman-constraint},\eqref{eq:pos-and-mu-state-npb}-\eqref{eq:convex-policy-constraint-trust-region}.\nonumber
\end{align}

\paragraph{Verification Step.} After each iteration, the linearization might be inaccurate, i.e, the resulting policy $\Tilde{\policy}$ and \emph{potentially inaccurate} visitation counts $\Tilde{\StateActionOccup}^\discount_{\Tilde{\policy}}, \Tilde{\StateOccup}^\discount_{\Tilde{\policy}}$ might not be feasible to the nonconvex policy constraint~\eqref{eq:policy-constraint}.
As a consequence of the potential infeasibility, the currently attained (linearized) optimal cost might significantly differ from the \emph{realized cost} by the feasible visiation counts for the $\Tilde{\policy}$.   
Additionally, existing SCP schemes linearizes the nonconvex problem around the previously inaccurate solutions for $\Tilde{\StateActionOccup}^\discount_{\Tilde{\policy}}$, and $\Tilde{\StateOccup}^\discount_{\Tilde{\policy}}$, further propagating the inaccuracy. 
The proposed \emph{verification step} solves these issues. 
Given the computed policy $\Tilde{\sigma}$, \texttt{SCPForward} computes the \emph{unique and sound} solution for the visitation count $\StateOccup^\discount_{\Tilde{\policy}}$ by solving the corresponding \emph{Bellman flow} constraints:
\begin{align}
    \StateOccup^\discount_{\Tilde{\policy}} (\state)  =& \Initdist(\state) + \label{eq:verif-bellman} \\
    &\discount \sum_{\state' \in \States} \sum_{\action \in \Actions} \Transition(\state | \state',\action) \StateOccup^\discount_{\Tilde{\policy}} (\state') \sum_{\observation \in \ObservationSet}\Observations(\observation|\state)\Tilde{\policy}_{\observation,\action}, \nonumber 
\end{align}%
for all $\state \in \States$, and where $\StateOccup^\discount_{\Tilde{\policy}} \geq 0$ is the only variable of the linear program. Then, \texttt{SCPForward} computes $\StateActionOccup^\discount_{\Tilde{\policy}}(\state,\action) = \StateOccup^\discount_{\Tilde{\policy}} (\state') \sum_{\observation \in \ObservationSet}\Observations(\observation|\state)\Tilde{\policy}_{\observation,\action}$ and the \emph{realized cost} cost at the current iteration is defined by%
\begin{align}
    \mathrm{C}(\Tilde{\sigma}, \Rewardvec^k) = & \quad \sum\nolimits_{(\state,\action)\in \States \times \Actions} -\log \frac{\StateActionOccup^\discount_{\Tilde{\policy}}(\state,\action)}{\StateOccup^\discount_{\Tilde{\policy}}} \StateActionOccup^\discount_{\Tilde{\policy}}(\state,\action) \nonumber\\
    &+  \sum\nolimits_{(\state,\action) \in \States \times \Actions} (\Rewardvec^k)^\mathrm{T} \phi(\state,\action) \StateActionOccup^\discount_{\Tilde{\policy}}(\state,\action),\label{eq:realized-cost}
\end{align}%
where we assume $0 \log 0 = 0$. Finally, if the realized cost $\mathrm{C}(\Tilde{\sigma}, \Rewardvec^k)$ does not improve over the previous cost $\mathrm{C}(\hat{\sigma}, \Rewardvec^k)$, the verification step rejects the obtained policy $\Tilde{\policy}$, contracts the trust region and $\texttt{SCPForward}$ iterates with the previous solutions $\hat{\policy}$, $\StateOccup^\discount_{\hat{\policy}}$, and $\StateActionOccup^\discount_{\hat{\policy}}$ . 
Otherwise, the linearization is sufficiently accurate, the trust region is expanded, and $\texttt{SCPForward}$ iterates with $\Tilde{\policy}$, $\StateOccup^\discount_{\Tilde{\policy}}$ and $\StateActionOccup^\discount_{\Tilde{\policy}}$.
\emph{By incorporating this verification step, we ensure that $\texttt{SCPForward}$ always linearizes the nonconvex optimization problem around a solution that satisfies the nonconvex constraint~\eqref{eq:policy-constraint}.}%

\algdef{SE}[DOWHILE]{Do}{doWhile}{\algorithmicdo}[1]{\algorithmicwhile\ #1}%
\begin{algorithm*}[t]
    \caption{\texttt{SCPForward:} Linear programming-based algorithm to solve the \emph{forward problem}~\eqref{eq:cost-entropy-npb}--\eqref{eq:pos-mu-state-npb}, i.e., compute a policy $\policy^k$ that maximizes the causal entropy, considers the feature matching, and satisfies the specifications, if available.}\label{algo:scpforward}
    \begin{algorithmic}[1]    
        \Require{Current weight estimate $\Rewardvec^k$, current best policy $\hat{\policy}$, side information $\varphi$ and $\lambda$, trust region $\trustregion > 1$, penalization coefficients $\beta$, $\beta^{\mathrm{sp}} \geq 0$, constant $\trustregion_0$ to expand or contract trust region, and a threshold $\trustregion_{\mathrm{lim}}$ for trust region contraction.}
        \State Find $\StateOccup^\discount_{\hat{\policy}}$ via linear constraint~\eqref{eq:verif-bellman} and $\StateActionOccup^\discount_{\hat{\policy}}=\StateOccup^\discount_{\hat{\policy}} (\state') \sum_{\observation \in \ObservationSet}\Observations(\observation|\state)\hat{\policy}_{\observation,\action}$, given $\hat{\policy}$ \Comment{Realized visitation counts} \label{alg:verif1}
        \State Find $\StateOccup^\mathrm{sp}_{\hat{\policy}}$ via linear constraint~\eqref{eq:ltl-bellman} with $\StateActionOccup^\mathrm{sp}_{\hat{\policy}}=\StateOccup^\mathrm{sp}_{\hat{\policy}} (\state') \sum_{\observation \in \ObservationSet}\Observations(\observation|\state)\hat{\policy}_{\observation,\action}$, given $\hat{\policy}$ \Comment{If $\reachPropSymbol$ is available} \label{alg:verif2}
        \State Compute the realized cost $\mathrm{C}(\hat{\sigma}, \Rewardvec^k) \gets \eqref{eq:realized-cost} + \mathrm{C}^{\mathrm{sp}}_{\hat{\policy}}$, given $\hat{\policy}$ \Comment{Add specifications' violation} \label{alg:verif3}
        \While{$\trustregion > \trustregion_{\mathrm{lim}}$} \Comment{Trust region threshold}
            \State Find optimal $\Tilde{\policy}$ to the augmented LP~\eqref{eq:cost-linearized-pb} via $\hat{\policy}$, $\StateOccup^\discount_{\hat{\policy}}$, $\StateActionOccup^\discount_{\hat{\policy}}$, $\StateOccup^\mathrm{sp}_{\hat{\policy}}$, $\StateActionOccup^\mathrm{sp}_{\hat{\policy}}$ \Comment{We augment the LP with constraints ~\eqref{eq:pos-and-mu-state-npb},~\eqref{eq:pos-mu-state-npb},~\eqref{eq:convex-policy-constraint-slack},~\eqref{eq:ltl-bellman}, and $(\mathrm{spec})$ induced by $\StateOccup^\mathrm{sp}_{\policy}, \StateActionOccup^\mathrm{sp}_{\policy}$, and by adding $-\beta \sum_{(\state,\action) \in \States\times\Actions}\slack^{\mathrm{sp}}_{\state,\action} -\beta^{\mathrm{sp}} \Gamma^\mathrm{sp}$ to the cost~\eqref{eq:cost-linearized-pb}.}
            \State Compute the realized $\StateOccup^\discount_{\Tilde{\policy}}$, $\StateActionOccup^\discount_{\Tilde{\policy}}$,$\StateOccup^\mathrm{sp}_{\Tilde{\policy}}$, $\StateActionOccup^\mathrm{sp}_{\Tilde{\policy}}$, and $\mathrm{C}(\Tilde{\sigma}, \Rewardvec^k)$ via $\Tilde{\policy}$ as in lines~\ref{alg:verif1}--\ref{alg:verif3} 
            \State \{$\hat{\policy} \gets \Tilde{\policy}$; $\trustregion \gets \trustregion \trustregion_0$\} if $\mathrm{C}(\Tilde{\sigma}, \Rewardvec^k) \geq \mathrm{C}(\hat{\sigma}, \Rewardvec^k)$ else \{$\trustregion \gets \trustregion / \trustregion_0$\} \Comment{Verification step}
        \EndWhile
    \State \Return $\policy^k := \hat{\policy}$
  \end{algorithmic}
\end{algorithm*}%

\subsection{Incorporating High-Level Task Specifications}
Given high-level side information on the agent tasks as the LTL formula $\reachPropSymbol$, we first compute the product of the POMDP and the $\omega$-automaton representing $\reachPropSymbol$ to find the set $\mathcal{T} \subseteq \States$ of states, called target or reach states, satisfying $\varphi$ with probability $1$ by using standard graph-based algorithms as a part of preprocessing step. We refer the reader to ~\citet{BK08} for a detailed introduction on how LTL specifications can be reduced to reachability specifications given by $\mathcal{T}$.

As a consequence, the probability of satisfying $\reachPropSymbol$ is the sum of the probability of reaching the target states $s \in \mathcal{T}$, which are given by the \emph{undiscounted state visitation count} $\StateOccup^\mathrm{sp}_{\policy}$. That is, $\mathrm{Pr}_{\Pomdp}^\policy(\varphi) = \sum_{s \in \mathcal{T}} \StateOccup^\mathrm{sp}_{\policy}(s)$. Unless $\discount = 1$, $\StateOccup^\mathrm{sp}_{\policy} \neq \StateOccup^\discount_{\policy}$. 
Thus, we introduce new variables $\StateOccup^\mathrm{sp}_{\policy}, \StateActionOccup^\mathrm{sp}_{\policy}$, and the adequate constraints in the linearized problem~\eqref{eq:cost-linearized-pb}.

\paragraph{Incorporating Undiscounted Visitation Variables to Linearized Problem.} We append new constraints, similar to~\eqref{eq:pos-and-mu-state-npb},~\eqref{eq:pos-mu-state-npb}, and~\eqref{eq:convex-policy-constraint-slack}, into the linearized problem~\eqref{eq:cost-linearized-pb}, where the variables $\StateOccup^\discount_{\policy}, \StateActionOccup^\discount_{\policy}, \slack_{\state,\action}, \StateOccup^\discount_{\hat{\policy}}$, $\StateActionOccup^\discount_{\hat{\policy}}$ are replaced by $\StateOccup^\mathrm{sp}_{\policy}, \StateActionOccup^\mathrm{sp}_{\policy}$, $\slack^{\mathrm{sp}}_{\state,\action}, \StateOccup^{\mathrm{sp}}_{\hat{\policy}}$, $\StateActionOccup^{\mathrm{sp}}_{\hat{\policy}}$, respectively. 
Further, we add the constraint
\begin{align}
    \StateOccup^{\mathrm{sp}}_{\policy} (\state)  = \Initdist(\state) + \sum_{\state' \in \States \setminus \mathcal{T}} \sum_{\action \in \Actions} \Transition(\state | \state',\action) \StateActionOccup^\mathrm{sp}_{\policy} (\state',\action), \label{eq:ltl-bellman}
\end{align}
which is a modification of the \emph{Bellman flow constraints} such that $\StateOccup^\mathrm{sp}_{\policy}(s)$ for all $s \in \mathcal{T}$ only counts transitions from non-target states. 
Finally, we penalize the introduced slack variables for feasibility of the linearization by augmenting the cost function with the term $-\beta \sum_{(\state,\action) \in \States\times\Actions}\slack^{\mathrm{sp}}_{\state,\action}$.

\paragraph{Relaxing Specification Constraints.} We add the constraint $(\mathrm{spec}) := \sum_{s \in \mathcal{T}} \StateOccup^\mathrm{sp}_{\policy}(s) + \Gamma^\mathrm{sp} \geq \lambda $ to the linearized problem, where $\Gamma^\mathrm{sp} \geq 0$ is a slack variable ensuring the linearized problem is always feasible. 
We augment the cost function with $-\beta^{\mathrm{sp}} \Gamma^\mathrm{sp}$ to penalize violating $\varphi$, where $\beta^{\mathrm{sp}}$ is a hyperparameter positive constant
.%

\paragraph{Updating Verification Step.} We modify the previously-introduced realized cost $\mathrm{C}(\Tilde{\sigma}, \Rewardvec^k)$ to penalize if the obtained policy does not satisfy the specification $\varphi$.
This cost also accounts for the linearization inaccuracy of the new policy constraint due to $\policy$, $\StateOccup^\mathrm{sp}_\policy$, and $\StateActionOccup^\mathrm{sp}_\policy$. 
At each iteration, $\texttt{SCPForward}$ computes the accurate $\StateOccup^\mathrm{sp}_{\Tilde{\policy}}$ of current policy $\Tilde{\policy}$ through solving a feasibility LP with constraints given by the \emph{modified Bellman flow constraints}~\eqref{eq:ltl-bellman}. 
Then, it augments $\mathrm{C}^{\mathrm{sp}}_{\Tilde{\policy}} = \min \{0, (\sum_{s \in \mathcal{T}} \StateOccup^\mathrm{sp}_{\Tilde{\policy}}(s) - \lambda) \beta^{\mathrm{sp}}\}$ to the realized cost to take the specification constraints into account.%
\section{Numerical Experiments}
We evaluate the proposed IRL algorithm on several POMDP instances, from \citet{junges2020enforcing}. We first compare our IRL algorithm with a straightforward variant of GAIL~\cite{ho2016generative} adapted for POMDPs. Then, we provide some results on the data-efficiency of the approach when taking advantage of side information. Finally, we demonstrate the scalability of the routine \texttt{SCPForward} for solving the \emph{forward} problem through comparisons with state-of-the-art solvers such as  $\texttt{SolvePOMDP}$~\cite{walraven2017accelerated}, $\texttt{SARSOP}$~\cite{kurniawati2008sarsop}, $\texttt{PRISM-POMDP}$~\cite{norman2017verification}. We consider throughout this section the hyperparameters $\beta = 1e^3$, $\beta^{\mathrm{sp}} = 10$, $\trustregion=1.01$, $\trustregion_0=1.5$, $\rho_{\mathrm{lim}}=1e^{-4}$, and $\discount=0.999$. Besides, we provide in the supplementary materials additional details on the key results of this paper and the experiments, e.g., preprocessing steps such as the product POMDP with $\mathrm{M}$-FSC or computing reachability specifications from LTL specifications.
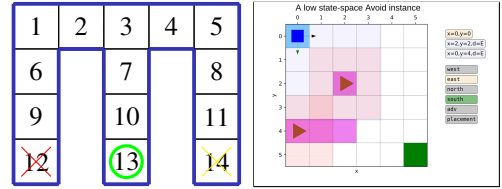
\begin{figure}[!hbt]
    \centering
    \definecolor{cheesebound}{rgb}{0.2, 0.2, 0.7}
\tikzset{cross/.style={cross out, draw, 
         minimum size=2*(#1-\pgflinewidth), 
         inner sep=0pt, outer sep=0pt}}
         
\begin{tikzpicture}[scale=0.6]
\draw[thin] (0,3) rectangle node {\color{black}1} (1,2);
\draw (1,3) rectangle node {\color{black}2} (2,2);
\draw (2,3) rectangle node {\color{black}3} (3,2);
\draw (3,3) rectangle node {\color{black}4} (4,2);
\draw (4,3) rectangle node {\color{black}5} (5,2);
\draw (0,2) rectangle node {\color{black}6} (1,1);
\draw (0,1) rectangle node {\color{black}9} (1,0);
\draw (0,0) rectangle node {\color{black}12} (1,-1); 
\draw (2,2) rectangle node {\color{black}7} (3,1);
\draw (2,1) rectangle node {\color{black}10} (3,0);
\draw (2,0) rectangle node {\color{black}13} (3,-1); 
\draw (4,2) rectangle node {\color{black}8} (5,1);
\draw (4,1) rectangle node {\color{black}11} (5,0);
\draw (4,0) rectangle node {\color{black}14} (5,-1); 

\draw (0.5, -0.5) node[cross=6pt, draw=red] {};
\draw [draw=green, very thick](2.5, -0.5) circle (10pt) {};
\draw (4.5, -0.5) node[cross=6pt, draw=yellow] {};

\draw[line width=0.5mm,cheesebound] (0,-1) edge (1,-1);
\draw[line width=0.5mm,cheesebound] (0,-1) edge (0,3);
\draw[line width=0.5mm,cheesebound] (5,3) edge (0,3);
\draw[line width=0.5mm,cheesebound] (1,2) edge (2,2);
\draw[line width=0.5mm,cheesebound] (1,2) edge (2,2);
\draw[line width=0.5mm,cheesebound] (3,2) edge (4,2);

\draw[line width=0.5mm,cheesebound] (2,-1) edge (3,-1);
\draw[line width=0.5mm,cheesebound] (4,-1) edge (5,-1);

\draw[line width=0.5mm,cheesebound] (5,-1) edge (5,3);

\draw[line width=0.5mm,cheesebound] (4,-1) edge (4,2);
\draw[line width=0.5mm,cheesebound] (3,-1) edge (3,2);
\draw[line width=0.5mm,cheesebound] (2,-1) edge (2,2);
\draw[line width=0.5mm,cheesebound] (1,-1) edge (1,2);

\end{tikzpicture}
    \frame{\includegraphics[scale=0.2]{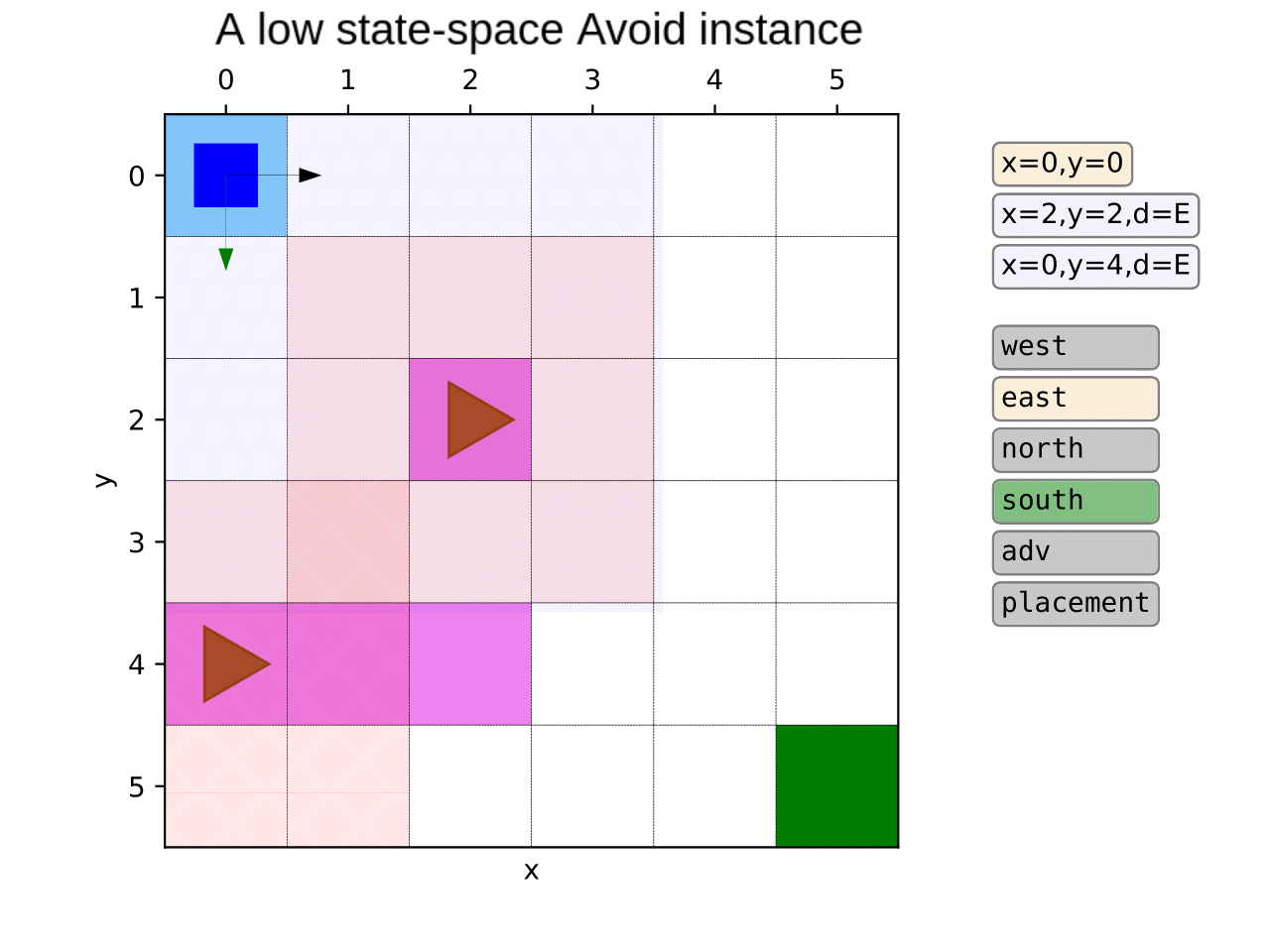}}
    \caption{Some examples from the benchmark set. From left to right, we have the \emph{Maze} and \emph{Avoid}, respectively.}
    \label{fig:benchmarkset}
\end{figure}
\begin{figure*}[t]
\centering
    \begin{subfigure}[t]{\textwidth}
        \centering\hspace*{2.4cm}\begin{tikzpicture}
\definecolor{color1}{HTML}{FF0000}
\definecolor{color0}{HTML}{0000FF}
\definecolor{color2}{HTML}{FF00FF}
\begin{customlegend}[legend columns=3,legend style={
  fill opacity=0.8,
  draw opacity=1,
  text opacity=1,
  at={(1.3,0.4)},
  anchor=south,
  draw=white!80!black,
  scale=0.40,
    font=\titlesize,
  mark options={scale=0.5},
},legend entries={No information asymmetry, Under information asymmetry, GAIL}]

\addlegendimage{line width=2pt, color0}
\addlegendimage{line width=2pt, color1, dashed}
\addlegendimage{line width=2pt, color2, densely dashdotted}
\end{customlegend}
\end{tikzpicture}
    \end{subfigure}\hfill%
    \centering
    \begin{subfigure}[t]{0.541\textwidth}
        \centering
        \centering\captionsetup{width=.95\linewidth}%
\begin{tikzpicture}

\definecolor{color1}{HTML}{FF0000}
\definecolor{color0}{HTML}{0000FF}
\definecolor{color2}{HTML}{FF00FF}

\begin{axis}[
legend cell align={left},
legend columns=2,
legend style={
  fill opacity=0.8,
  draw opacity=1,
  text opacity=1,
  at={(1.15,1.25)},
  anchor=south,
    ticklabel style = {font=\tiny},
  draw=white!80!black,
  font=\fontsize{7.4}{7.4}\selectfont,
  scale=0.40
},
title={$R_\policy^\phi$},
title style={yshift=-1.5ex,xshift=-15ex,font=\titlesize},
width=6.5cm,
height=3.1cm,
tick align=outside,
tick pos=left,
no markers,
every axis plot/.append style={line width=2pt},
x grid style={white!69.0196078431373!black},
xlabel={\footnotesize{Time steps}},
xlabel={\textcolor{gray!40!black}{\arial \textit{Finite-memory} policy}},
xlabel style={font=\titlesize,yshift=18ex},
xmajorgrids,
xmin=-4.95, xmax=103.95,
xtick style={color=black},
y grid style={white!69.0196078431373!black},
ylabel={\footnotesize{Mean  accumulated  reward}},
ylabel={\textcolor{gray!40!black}{\arial\textbf{Without} side} \\\textcolor{gray!40!black}{\arial information}},
ylabel style={rotate=-90,align=center,font=\titlesize},
xtick={0,25,50,75,100},
ymajorgrids,
ymin=-30.8979836098955, ymax=66.0315604206547,
ytick style={color=black}
]

\addplot[color0] table[x=x,y=y,mark=none] {maze_sim_files/maze_mem10_trajsize5pomdp_irl.tex};
\addplot[color1,dashed] table[x=x,y=y,mark=none] {maze_sim_files/maze_mem10_trajsize5mdp_irl.tex};


\addplot[color2,densely dashdotted] table[x=x,y=y,mark=none] {maze_sim_files/maze_mdp_fwd_gail.tex};
\addplot [name path=upper2,draw=none] table[x=x,y expr=\thisrow{y}+\thisrow{err}] {maze_sim_files/maze_mdp_fwd_gail.tex};
\addplot [name path=lower2,draw=none] table[x=x,y expr=\thisrow{y}-\thisrow{err}] {maze_sim_files/maze_mdp_fwd_gail.tex};
\addplot [fill=color2!40,opacity=0.5] fill between[of=upper2 and lower2];

\addplot [name path=upper1,draw=none] table[x=x,y expr=\thisrow{y}+\thisrow{err}]{maze_sim_files/maze_mem10_trajsize5mdp_irl.tex};
\addplot [name path=lower1,draw=none] table[x=x,y expr=\thisrow{y}-\thisrow{err}] {maze_sim_files/maze_mem10_trajsize5mdp_irl.tex};
\addplot [fill=color1!40,opacity=0.5] fill between[of=upper1 and lower1];

\addplot [name path=upper,draw=none] table[x=x,y expr=\thisrow{y}+\thisrow{err}] {maze_sim_files/maze_mem10_trajsize5pomdp_irl.tex};
\addplot [name path=lower,draw=none] table[x=x,y expr=\thisrow{y}-\thisrow{err}] {maze_sim_files/maze_mem10_trajsize5pomdp_irl.tex};
\addplot [fill=color0!40,opacity=0.5] fill between[of=upper and lower];

\end{axis}

\end{tikzpicture}
    \end{subfigure}%
    \begin{subfigure}[t]{0.459\textwidth}
        \centering
        \centering\captionsetup{width=.95\linewidth}%
\begin{tikzpicture}

\definecolor{color1}{HTML}{FF0000}
\definecolor{color0}{HTML}{0000FF}
\definecolor{color2}{HTML}{FF00FF}

\begin{axis}[
legend cell align={left},
legend columns=2,
legend style={
  fill opacity=0.8,
  draw opacity=1,
  text opacity=1,
  at={(0.5,1)},
  anchor=south,
  ticklabel style = {font=\tiny},
  draw=white!80!black,
  font=\fontsize{6}{6}\selectfont,
  scale=0.40
},
title={$R_\policy^\phi$},
title style={yshift=-1.5ex,xshift=-15ex,font=\titlesize},
width=6.5cm,
height=3.1cm,
tick align=outside,
tick pos=left,
no markers,
every axis plot/.append style={line width=2pt},
x grid style={white!69.0196078431373!black},
xlabel={\footnotesize{Time steps}},
xlabel={},
xlabel={\textcolor{gray!40!black}{\arial \textit{Memoryless} policy}},
xlabel style={font=\titlesize,yshift=18ex},
xtick={0,25,50,75,100},
xmajorgrids,
xmin=-4.95, xmax=103.95,
xtick style={color=black},
y grid style={white!69.0196078431373!black},
ylabel={\footnotesize{Mean  accumulated  reward}},
ylabel={},
ymajorgrids,
ymin=-30.8979836098955, ymax=66.0315604206547,
ytick style={color=black}
]

\addplot[color0] table[x=x,y=y,mark=none] {maze_sim_files/maze_mem1_trajsize5pomdp_irl.tex};
\addplot[color1,dashed] table[x=x,y=y,mark=none] {maze_sim_files/maze_mem1_trajsize5mdp_irl.tex};


\addplot[color2,densely dashdotted] table[x=x,y=y,mark=none] {maze_sim_files/maze_mdp_fwd_gail.tex};
\addplot [name path=upper2,draw=none] table[x=x,y expr=\thisrow{y}+\thisrow{err}] {maze_sim_files/maze_mdp_fwd_gail.tex};
\addplot [name path=lower2,draw=none] table[x=x,y expr=\thisrow{y}-\thisrow{err}] {maze_sim_files/maze_mdp_fwd_gail.tex};
\addplot [fill=color2!40,opacity=0.5] fill between[of=upper2 and lower2];

\addplot [name path=upper1,draw=none] table[x=x,y expr=\thisrow{y}+\thisrow{err}]{maze_sim_files/maze_mem1_trajsize5mdp_irl.tex};
\addplot [name path=lower1,draw=none] table[x=x,y expr=\thisrow{y}-\thisrow{err}] {maze_sim_files/maze_mem1_trajsize5mdp_irl.tex};
\addplot [fill=color1!40,opacity=0.5] fill between[of=upper1 and lower1];

\addplot [name path=upper,draw=none] table[x=x,y expr=\thisrow{y}+\thisrow{err}] {maze_sim_files/maze_mem1_trajsize5pomdp_irl.tex};
\addplot [name path=lower,draw=none] table[x=x,y expr=\thisrow{y}-\thisrow{err}] {maze_sim_files/maze_mem1_trajsize5pomdp_irl.tex};
\addplot [fill=color0!40,opacity=0.5] fill between[of=upper and lower];

\end{axis}

\end{tikzpicture}
    \end{subfigure}%
    \vfill
    \centering
        \begin{subfigure}[t]{0.541\textwidth}
        \centering
        \centering\captionsetup{width=.95\linewidth}%
\begin{tikzpicture}

\definecolor{color1}{HTML}{FF0000}
\definecolor{color0}{HTML}{0000FF}
\definecolor{color2}{HTML}{FF00FF}
\begin{axis}[
legend cell align={left},
legend columns=2,
legend style={
  fill opacity=0.8,
  draw opacity=1,
  text opacity=1,
  at={(0.5,1)},
  anchor=south,
  ticklabel style = {font=\tiny},
  draw=white!80!black,
  font=\fontsize{6}{6}\selectfont,
  scale=0.40
},
title={\textbf{Finite-memory} policy \emph{with} side information},
title={$R_\policy^\phi$},
title style={yshift=-1.5ex,xshift=-15ex,font=\titlesize},
width=6.5cm,
height=3.1cm,
tick align=outside,
tick pos=left,
no markers,
every axis plot/.append style={line width=2pt},
x grid style={white!69.0196078431373!black},
xlabel={\titlesize{Time Steps}},
xmajorgrids,
xmin=-4.95, xmax=103.95,
xtick style={color=black},
y grid style={white!69.0196078431373!black},
xtick={0,25,50,75,100},
ylabel={\footnotesize{Mean accumulated  reward}},
ylabel={\titlesize{$R_\policy^\phi$}},
ylabel={\textcolor{gray!40!black}{\;\;\,\arial\textbf{With} side\;\;\,} \\\textcolor{gray!40!black}{\arial information}},
ylabel style={rotate=-90,align=center,font=\titlesize},
ymajorgrids,
ymin=-30.8979836098955, ymax=66.0315604206547,
ytick style={color=black}
]

\addplot[color0] table[x=x,y=y,mark=none] {maze_sim_files/maze_mem10_trajsize5pomdp_irl_si.tex};
\addplot[color1,dashed] table[x=x,y=y,mark=none] {maze_sim_files/maze_mem10_trajsize5mdp_irl_si.tex};


\addplot[color2,densely dashdotted] table[x=x,y=y,mark=none] {maze_sim_files/maze_mdp_fwd_gail.tex};
\addplot [name path=upper2,draw=none] table[x=x,y expr=\thisrow{y}+\thisrow{err}] {maze_sim_files/maze_mdp_fwd_gail.tex};
\addplot [name path=lower2,draw=none] table[x=x,y expr=\thisrow{y}-\thisrow{err}] {maze_sim_files/maze_mdp_fwd_gail.tex};
\addplot [fill=color2!40,opacity=0.5] fill between[of=upper2 and lower2];

\addplot [name path=upper1,draw=none] table[x=x,y expr=\thisrow{y}+\thisrow{err}]{maze_sim_files/maze_mem10_trajsize5mdp_irl_si.tex};
\addplot [name path=lower1,draw=none] table[x=x,y expr=\thisrow{y}-\thisrow{err}] {maze_sim_files/maze_mem10_trajsize5mdp_irl_si.tex};
\addplot [fill=color1!40,opacity=0.5] fill between[of=upper1 and lower1];

\addplot [name path=upper,draw=none] table[x=x,y expr=\thisrow{y}+\thisrow{err}] {maze_sim_files/maze_mem10_trajsize5pomdp_irl_si.tex};
\addplot [name path=lower,draw=none] table[x=x,y expr=\thisrow{y}-\thisrow{err}] {maze_sim_files/maze_mem10_trajsize5pomdp_irl_si.tex};
\addplot [fill=color0!40,opacity=0.5] fill between[of=upper and lower];

\end{axis}

\end{tikzpicture}
    \end{subfigure}%
    \begin{subfigure}[t]{0.459\textwidth}
        \centering
        \centering\captionsetup{width=.95\linewidth}%
\begin{tikzpicture}

\definecolor{color1}{HTML}{FF0000}
\definecolor{color0}{HTML}{0000FF}
\definecolor{color2}{HTML}{FF00FF}
\begin{axis}[
legend cell align={left},
legend columns=2,
legend style={
  fill opacity=0.8,
  draw opacity=1,
  text opacity=1,
  at={(0.5,1)},
  anchor=south,
  ticklabel style = {font=\tiny},
  draw=white!80!black,
  font=\fontsize{6}{6}\selectfont,
  scale=0.40
},
title={$R_\policy^\phi$},
title style={yshift=-1.5ex,xshift=-15ex,font=\titlesize},
width=6.5cm,
height=3.1cm,
tick align=outside,
tick pos=left,
no markers,
every axis plot/.append style={line width=2pt},
x grid style={white!69.0196078431373!black},
xlabel={\titlesize{Time Steps}},
xmajorgrids,
xmin=-4.95, xmax=103.95,
xtick style={color=black},
y grid style={white!69.0196078431373!black},
ylabel={\footnotesize{Mean  accumulated  reward}},
ylabel={},
ymajorgrids,
xtick={0,25,50,75,100},
ymin=-30.8979836098955, ymax=66.0315604206547,
ytick style={color=black}
]

\addplot[color0] table[x=x,y=y,mark=none] {maze_sim_files/maze_mem1_trajsize5pomdp_irl_si.tex};
\addplot[color1,dashed] table[x=x,y=y,mark=none] {maze_sim_files/maze_mem1_trajsize5mdp_irl_si.tex};


\addplot[color2,densely dashdotted] table[x=x,y=y,mark=none] {maze_sim_files/maze_mdp_fwd_gail.tex};
\addplot [name path=upper2,draw=none] table[x=x,y expr=\thisrow{y}+\thisrow{err}] {maze_sim_files/maze_mdp_fwd_gail.tex};
\addplot [name path=lower2,draw=none] table[x=x,y expr=\thisrow{y}-\thisrow{err}] {maze_sim_files/maze_mdp_fwd_gail.tex};
\addplot [fill=color2!40,opacity=0.5] fill between[of=upper2 and lower2];

\addplot [name path=upper1,draw=none] table[x=x,y expr=\thisrow{y}+\thisrow{err}]{maze_sim_files/maze_mem1_trajsize5mdp_irl_si.tex};
\addplot [name path=lower1,draw=none] table[x=x,y expr=\thisrow{y}-\thisrow{err}] {maze_sim_files/maze_mem1_trajsize5mdp_irl_si.tex};
\addplot [fill=color1!40,opacity=0.5] fill between[of=upper1 and lower1];

\addplot [name path=upper,draw=none] table[x=x,y expr=\thisrow{y}+\thisrow{err}] {maze_sim_files/maze_mem1_trajsize5pomdp_irl_si.tex};
\addplot [name path=lower,draw=none] table[x=x,y expr=\thisrow{y}-\thisrow{err}] {maze_sim_files/maze_mem1_trajsize5pomdp_irl_si.tex};
\addplot [fill=color0!40,opacity=0.5] fill between[of=upper and lower];

\end{axis}

\end{tikzpicture}       
    \end{subfigure}%
    \caption{Representative results on the $\mathrm{Maze}$ example showing the reward of the policies under the true reward function ($R_\policy^\phi$) versus the time steps.
  Compare the two rows: The policies in the top row that do not utilize side information suffer a performance drop under information asymmetry.
    On the other hand, in the bottom row, the performance of policies incorporating side information into learning does not decrease under information asymmetry.
    Compare the two columns: The performance of the finite-memory policies in the left column is significantly better than memoryless policies. Except for the memoryless policies without side information, our algorithm outperforms GAIL. The expert reward on the MDP is $48.22$, while $47.83$ on POMDP.}
    \label{fig:maze}
\end{figure*}
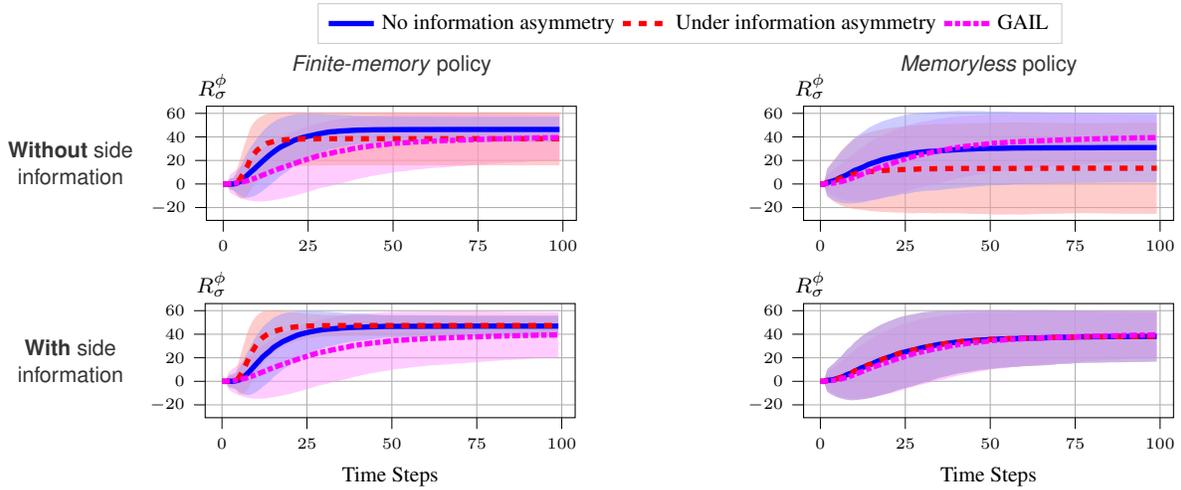%

\paragraph{Benchmark Set.} The POMDP instances are as follows. $\emph{Evade}$ is a turn-based game where the agent must reach a destination without being intercepted by a faster player. 
In $\emph{Avoid}$, the agent must avoid being detected by two other moving players following certain preset, yet unknown routes.
In $\emph{Intercept}$, the agent must intercept another player who is trying to exit a gridworld. 
In $\emph{Rocks}$, the agents must sample at least one good rock over the several rocks without any failures. 
Finally, in $\emph{Maze}$, the agent must exit a maze as fast as possible while avoiding trap states.

\paragraph{Variants of Learned Policies and Experts.}
We refer to four types of policies.
The type of policy depends on whether it uses side information from a temporal specification $\reachPropSymbol$ or not, and whether it uses a memory size $\mathrm{M}=1$ or $\mathrm{M}=10$.
We also consider two types of experts.
The first expert has full information about the environment and computes an optimal policy in the underlying MDP.
The second expert has partial observation and computes a locally optimal policy in the POMDP with a memory size of $\mathrm{M}=15$.
Recall that the agent always has partial information.
Therefore, the first type of expert corresponds to having information asymmetry between the learning agent and expert. \emph{Besides, we consider as a baseline a variant of GAIL where we learn the policy on the MDP without side information, and extend it to POMDPs via an offline computation of the belief in the states. Doing so provides a significant advantage to GAIL since we learn on the MDP. We do not compare with~\citet{choi2011inverse} as explained in the related work.}

We discuss the effect of side information and memory in the policies.
While we detail only on the \emph{Maze} example, where the agent must exit a maze as fast as possible, we observe similar patterns for other examples.
We give detailed results for the other examples in the supplementary material.

\subsection{Maze Example}

The POMDP $\Pomdp$ is specified by $\States = \{s_1,\hdots,s_{14}\}$ corresponding to the cell labels in Figure~\ref{fig:benchmarkset}. An agent in the maze only observes whether or not there is a wall (in blue) in a neighboring cell. That is, the set of observations is $\Observations = \{o_1,\hdots,o_6, o_7\}$. For example, $o_1$ corresponds to observing west and north walls ($s_1$), $o_2$ to north and south walls ($s_2$, $s_4$), and $o_5$ to east and west walls ($s_6,s_7,s_8,s_9,s_{10},s_{11}$). The observations $o_6$ and $o_7$ denote the target state ($s_{13}$) and bad states($s_{12}$, $s_{14}$). The transition model is stochastic with a probability of slipping $p=0.1$. Further, the states $s_{13}$ and $s_{14}$ lead to the end of the simulation (trapping states).

In the IRL experiments, we consider three feature functions. We penalize taking more steps with $\phi^{\mathrm{time}}(\state,\action) = -1$ for all $\state,\action$. We provide a positive reward when reaching $s_{13}$ with $\phi^{\mathrm{target}}(\state,\alpha) = 1$ if $s = s_{13}$ and $\phi^{\mathrm{target}}(\state,\alpha) = 0$ otherwise. We penalize bad states $s_{12}$ and $s_{14}$ with $\phi^{\mathrm{bad}}(\state,\alpha) = -1$ if $s = s_{12}$ or $s = s_{14}$, and $\phi^{\mathrm{bad}}(\state,\alpha) = 0$ otherwise. \emph{Finally, we have the LTL formula $\reachPropSymbol = \textbf{G} \; \lnot \: \mathrm{bad}$ as the task specification, where $\mathrm{bad}$ is an atomic proposition that is true if the current state $\state = s_{12}$ or $\state = s_{14}$. We constrain the learned policy to satisfy $\mathrm{Pr}_{\Pomdp}^\policy(\textbf{G} \; \lnot \: \mathrm{bad}) \geq 0.9$.}

\paragraph{Side Information Alleviates the Information Asymmetry.} Figure~\ref{fig:maze} shows that if there is an information asymmetry between the learning agent and the expert, the policies that do not utilize side information suffer a significant performance drop.
The policies that do not incorporate side information into learning obtain a lower performance by $57$\% under information asymmetry, as shown in the top row of Figure~\ref{fig:maze}.
On the other hand, as seen in the bottom row of Figure~\ref{fig:maze}, the performance of the policies that use side information is almost unaffected by the information asymmetry.

\begin{table*}[t]
	\setlength{\tabcolsep}{4.5pt}
	\centering
	\scalebox{0.75}{%
		\begin{tabular}{@{}cccc|cc|cc|cc@{}}
			\toprule
			&  & & & \multicolumn{2}{c|}{$\texttt{SCPForward}$}   &\multicolumn{2}{c|}{$\texttt{SARSOP}$} & \multicolumn{2}{c}{$\texttt{SolvePOMDP}$}  \\
			Problem   & $|\States|$ & $|\States \times \Observations|$ & $|\Observations|$ & $R_\policy^\phi$ & Time (s) & $R_\policy^\phi$ & Time (s) & $R_\policy^\phi$ & Time (s) \\
			\midrule
			$\mathrm{Maze}$ & $17$  & $162$ & $11$ & $39.24$ &   $\mathbf{0.1}$ & $\mathbf{47.83}$ & $0.24$ &  $47.83$ & $0.33$\\
			$\mathrm{Maze}$($10$-$\mathrm{FSC})$ & $161$  & $2891$ & $101$ & $46.32$ &   $2.04$ & NA & NA &  NA & NA\\
			$\mathrm{Rock}$ & $550$  & $4643$ & $67$ & $19.68$ &   $12.2$ & $\mathbf{19.83}$ & $\mathbf{0.05}$ &  $-$ & $-$\\
			$\mathrm{Rock}$($5$-$\mathrm{FSC})$ & $2746$  & $41759$ & $331$ & $19.82$ &   $97.84$ & NA & NA &  NA & NA\\
			$\mathrm{Intercept}$ & $1321$  & $5021$ & $1025$ & $\mathbf{19.83}$ &   $\mathbf{10.28}$ & $\mathbf{19.83}$ & $13.71$ &  $-$ & $-$\\
			$\mathrm{Intercept}$ & $1321$  & $7041$ & $1025$ & $\mathbf{19.81}$ &   $\mathbf{13.18}$ & $\mathbf{19.81}$ & $81.19$ &  $-$ & $-$\\
			$\mathrm{Evade}$ & $2081$  & $16761$ & $1089$ & $\mathbf{96.79}$ &   $\mathbf{26.25}$ & $95.28$ & $3600$ &  $-$ & $-$\\
			$\mathrm{Evade}$ & $36361$  & $341121$ & $18383$ & $\mathbf{94.97}$ &   $\mathbf{3600}$ & $-$ & $-$ &  $-$ & $-$\\
			$\mathrm{Avoid}$ & $2241$  & $8833$ & $1956$ & $\mathbf{9.86}$ &   $\mathbf{14.63}$ & $\mathbf{9.86}$ & $210.47$ &  $-$ & $-$\\
			$\mathrm{Avoid}$ & $19797$  & $62133$ & $3164$ & $\mathbf{9.72}$ &   $\mathbf{3503}$ & $-$ & $-$ &  $-$ & $-$\\
			\bottomrule
	\end{tabular}}
	\caption{Results for the benchmarks. On larger benchmarks (e.g., $\mathrm{Evade}$ and $\mathrm{Avoid}$), the method we developed can compute a locally optimal policy. We set the time-out to $3600$ seconds. An empty cell (denoted by $-$) represents the solver failed to compute any policy before the time-out, while NA refers to not applicable due to the approach being based on belief updates.}
	\label{tab:ToolComp}
\end{table*}

\paragraph{Memory Leads to More Performant Policies.}
The results in Figure~\ref{fig:maze} demonstrate that incorporating memory into the policies improves the performance, i.e., the attained reward, in all examples, both in solving the forward problem and learning policies from expert demonstrations.
Incorporating memory partially alleviates the effects of  information asymmetry, as the performance of the finite-memory policy decreases by $18$\% under information asymmetry as opposed to $57$\% for the memoryless policy.

We see that in Table~\ref{tab:ToolComp}, incorporating memory into policy on the $\mathrm{Maze}$ and $\mathrm{Rocks}$ benchmarks, allows $\texttt{SCPForward}$ to compute policies that are almost optimal, evidenced by obtaining almost the same reward as the solver $\texttt{SARSOP}$.

\begin{figure}[!hbt]
    \centering
\begin{tikzpicture}

\begin{groupplot}[group style={group size=2 by 1}]
\nextgroupplot[
legend cell align={left},
legend columns=5,
legend style={
  fill opacity=0.8,
  draw opacity=1,
  text opacity=1,
  at={(-0.45,1.05)},
  anchor=south west,
  draw=white!80!black
},
width=4.5cm,
height=3.5cm,
tick align=outside,
tick pos=left,
x grid style={white!69.0196078431373!black},
xlabel={Number of trajectories},
xmajorgrids,
xmin=0.3, xmax=15.7,
xtick style={color=black},
y grid style={white!69.0196078431373!black},
ylabel={Total reward},
ymajorgrids,
ymin=12.15607, ymax=48.82853,
ytick style={color=black}
]
\addplot [semithick, blue, mark=asterisk, mark size=1, mark options={solid}]
table {%
1 13.8229999542236
2 15.6123332977295
3 21.7000007629395
4 28.0136661529541
5 30.5090007781982
6 35.466667175293
7 35.1426658630371
8 37.1469993591309
9 37.0320014953613
10 37.8953323364258
11 37.1166648864746
12 37.4440002441406
13 37.3230018615723
14 37.8726673126221
15 37.2173347473145
};
\addlegendentry{Without LTL}
\addplot [semithick, green!50.1960784313725!black, dashed, mark=asterisk, mark size=1, mark options={solid}]
table {%
1 30.3713340759277
2 33.8123321533203
3 36.5169982910156
4 37.4993324279785
5 37.666332244873
6 38.3046684265137
7 39.0913314819336
8 39.9926681518555
9 39.4426651000977
10 39.8916664123535
11 39.4599990844727
12 39.9243316650391
13 39.7676651000977
14 39.7566673278809
15 39.9330009460449
};
\addlegendentry{With LTL}
\addplot [semithick, red]
table {%
1 47.1615982055664
15 47.1615982055664
};
\addlegendentry{Opt. Rew. POMDP}

\nextgroupplot[
legend cell align={left},
legend columns=5,
legend style={
  fill opacity=0.8,
  draw opacity=1,
  text opacity=1,
  at={(0,1)},
  anchor=south west,
  draw=white!80!black
},
width=4.5cm,
height=3.5cm,
tick align=outside,
tick pos=left,
x grid style={white!69.0196078431373!black},
xlabel={Number of trajectories},
xmajorgrids,
xmin=0.3, xmax=15.7,
xtick style={color=black},
y grid style={white!69.0196078431373!black},
ymajorgrids,
ymin=38.59682, ymax=47.5694466666667,
ytick style={color=black}
]
\addplot [semithick, blue, mark=asterisk, mark size=1, mark options={solid}]
table {%
1 39.764331817627
2 39.564331817627
3 39.964331817627
4 39.9476661682129
5 40.0233345031738
6 40.6316680908203
7 40.8110008239746
8 39.964331817627
9 41.3343315124512
10 41.2949989318848
11 41.0046653747559
12 41.5623321533203
13 41.224666595459
14 41.5193328857422
15 41.4710006713867
};
\addplot [semithick, green!50.1960784313725!black, dashed, mark=asterisk, mark size=1, mark options={solid}]
table {%
1 40.7490005493164
2 41.2799987792969
3 41.099666595459
4 42.2910003662109
5 42.798999786377
6 42.7073348999023
7 42.7443321228027
8 42.601001739502
9 42.8359985351562
10 42.842000579834
11 42.8273323059082
12 42.8623344421387
13 43.02667388916
14 43.1749992370605
15 43.151668548584
};
\addplot [semithick, red, forget plot]
table {%
1 47.1615982055664
15 47.1615982055664
};
\end{groupplot}

\end{tikzpicture}
    \caption{We show the data efficiency of the proposed approach through the total reward obtained by the learned policies as a function of the number of expert demonstrations (No information asymmetry). The figure on the left shows the performance of learning memoryless policies, while the figure on the right shows the performance of a $5$-FSC.}
    \label{fig:data-eff}
\end{figure}
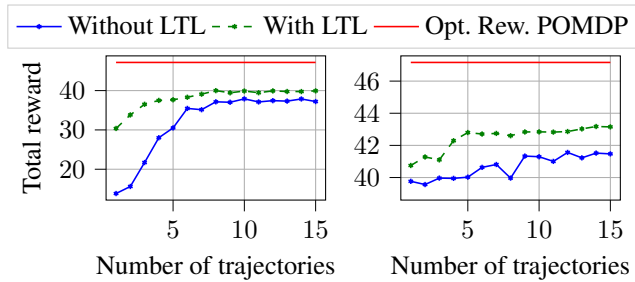

\paragraph{Side Information Improves Data Efficiency and Performance.}
Figure~\ref{fig:data-eff} shows that even on a low data regime, learning with task specifications achieves significantly better performance than without the task specifications. 

Besides, in a more complicated environment such as $\mathrm{Avoid}$, Figure~\ref{fig:avoid} shows that task specifications are crucial to hope even to learn the task. We refer the reader to the supplementary material for the details of the experiment.
\begin{figure}[!hbt]
\centering
    \centering
    \begin{subfigure}[t]{0.459\textwidth}
        \centering
        \centering\begin{tikzpicture}
\definecolor{color1}{HTML}{FF0000}
\definecolor{color0}{HTML}{0000FF}
\begin{customlegend}[legend columns=2,legend style={
  fill opacity=0.8,
  draw opacity=1,
  text opacity=1,
  at={(0.0,110.2)},
  anchor=south,
  draw=white!80!black,
  scale=0.40,
  mark options={scale=0.5},
},legend entries={With side information, Without side information}]
\addlegendimage{line width=2pt, color0}
\addlegendimage{line width=2pt, color1,dashed}
\end{customlegend}
\end{tikzpicture}
        \centering\captionsetup{width=.95\linewidth}%
\begin{tikzpicture}

\definecolor{color0}{HTML}{FF0000}
\definecolor{color1}{HTML}{0000FF}
\begin{axis}[
legend cell align={left},
legend columns=2,
legend style={
  fill opacity=0.8,
  draw opacity=1,
  text opacity=1,
  at={(0.5,1)},
  anchor=south,
  draw=white!80!black,
  font=\fontsize{6}{6}\selectfont,
  scale=0.40
},
title={$R_\policy^\phi$},
title={},
title style={yshift=-1.5ex,xshift=-15ex,font=\titlesize},
width=\textwidth,
height=0.4\textwidth,
tick align=outside,
tick pos=left,
no markers,
every axis plot/.append style={line width=2pt},
x grid style={white!69.0196078431373!black},
xlabel={{Time Steps}},
xmajorgrids,
xmin=-4.95, xmax=303.95,
xtick style={color=black},
y grid style={white!69.0196078431373!black},
ylabel={{$R_\policy^\phi$}},
ymajorgrids,
xtick={0,75,150,225,300},
ymin=-32.8979836098955, ymax=56.0315604206547,
ticklabel style = {font=\normalsize}
]

\addplot[color0,dashed] table[x=x,y=y,mark=none] {avoid_sim_files/avoid_mem1_trajsize10mdp_irl.tex};
\addplot[color1] table[x=x,y=y,mark=none] {avoid_sim_files/avoid_mem1_trajsize10mdp_irl_si.tex};

\addplot [name path=upper1,draw=none] table[x=x,y expr=\thisrow{y}+\thisrow{err}]{avoid_sim_files/avoid_mem1_trajsize10mdp_irl_si.tex};
\addplot [name path=lower1,draw=none] table[x=x,y expr=\thisrow{y}-\thisrow{err}] {avoid_sim_files/avoid_mem1_trajsize10mdp_irl_si.tex};
\addplot [fill=color1!40,opacity=0.5] fill between[of=upper1 and lower1];

\addplot [name path=upper,draw=none] table[x=x,y expr=\thisrow{y}+\thisrow{err}] {avoid_sim_files/avoid_mem1_trajsize10mdp_irl.tex};
\addplot [name path=lower,draw=none] table[x=x,y expr=\thisrow{y}-\thisrow{err}] {avoid_sim_files/avoid_mem1_trajsize10mdp_irl.tex};
\addplot [fill=color0!40,opacity=0.5] fill between[of=upper and lower];
\end{axis}

\end{tikzpicture}
    \end{subfigure}%
    \caption{Results on the $\mathrm{Avoid}$ example show that side information can help to crucially improve the performance.}
    \label{fig:avoid}
\end{figure}
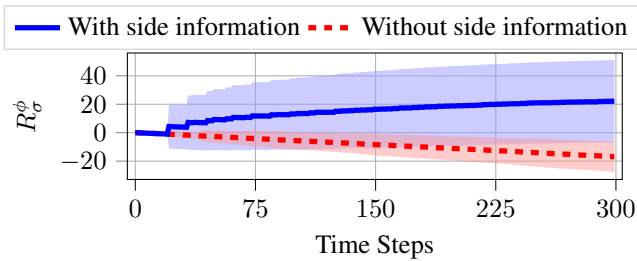

\subsection{\texttt{SCPForward} Yields Better Scalability}
We highlight three observations regarding the scalability of $\texttt{SCPForward}$.
First, the results in Table~\ref{tab:ToolComp} show that only $\texttt{SARSOP}$ is competitive with $\texttt{SCPForward}$ on larger POMDPs.
$\texttt{SolvePOMDP}$ runs out of time in all but the smallest benchmarks, and $\texttt{PrismPOMDP}$ runs out of memory in all benchmarks. Most of these approaches are based on updating a belief over the states, which for a large state space can become extremely computationally expensive.

Second, in the benchmarks with smaller state spaces, e.g., \emph{Maze} and \emph{Rock}, $\texttt{SARSOP}$ can compute policies that yield better performance in less time. This is due to the efficiency of belief-based approaches on small-size problems.
On the other hand, $\texttt{SARSOP}$ does not scale to larger POMDPs with a larger number of states and observations.
For example, by increasing the number of transitions in \emph{Intercept} benchmark from $5021$ to $7041$, the computation time for $\texttt{SARSOP}$ increases by $516$\%.
On the other hand, the increase of the computation time of $\texttt{SCPForward}$ is only $28$\%.

Third, on the largest benchmarks, including tens of thousands of states and observations, $\texttt{SARSOP}$ fails to compute any policy before time-out, while $\texttt{SCPForward}$ found a solution.
Finally, we also note that $\texttt{SCPForward}$ can also compute a policy that maximizes the causal entropy and satisfies an LTL specification, unlike $\texttt{SARSOP}$.
\section{Conclusion}
We develop an algorithm for inverse reinforcement learning under partial observation.
We empirically demonstrate that by incorporating task specifications into the learning process, we can alleviate the information asymmetry between the expert and the agent and learn policies that yield similar performance to the expert. Further, we show that our routine $\texttt{SCPForward}$ to solve the forward problem outperforms state-of-the-art POMDP solvers on instances with a large number of states and observations.

\paragraph{Work Limitations.} This work assumes that the transition and observation functions of the POMDP are known to the algorithm. Future work will investigate removing this assumption and developing model-free-based approaches. We will also integrate the framework with more expressive neural-network-based reward functions.

\newpage
\bibliography{ref.bib}

\newpage

\onecolumn

\begin{center} 
    \begin{LARGE}
        \textbf{Task-Guided Inverse Reinforcement Learning Under Partial Information\\
$-$ \Large \normalfont Supplementary Material $-$}
    \end{LARGE}
\end{center}

\section{Derivations of the Constraints and Incorporating Nonlinear Reward Functions into the Learning Algorithm}

In this section, we first recall the obtained expression of the causal entropy $H^\gamma_\policy$ as a function of the visitation counts $\StateOccup^\discount_\policy$ and $\StateActionOccup^\discount_\policy$.
We then prove the concavity of the causal entropy, which enables convex-optimization-based formulation of the task-guided inverse reinforcement learning (IRL) problem. 
Then, we provide additional details on the derivation of the affine constraint implied by the \emph{Bellman flow constraint}.
Finally, we show how the algorithm developed in this paper can be adapted to learn a reward function with a nonlinear parameterization, e.g., a neural network, of the observations.

\paragraph{Concave Causal Entropy. }
We first recall the definitions of the state and state-action visitation counts.
For a policy $\policy$, state $\state$, and action $\action$, the discounted state visitation counts are defined by $\StateOccup^\discount_\policy (\state) \triangleq \mathbb{E}_{S_t} [\sum_{t=1}^\infty \discount^t \mathbbm{1}_{\{S_t = \state\}}]$ and the discounted state-action visitation counts are defined by $\StateActionOccup^\discount_\policy (\state,\action) \triangleq \mathbb{E}_{A_t, S_t} [\sum_{t=1}^\infty \discount^t \mathbbm{1}_{\{S_t = \state, A_t = \action\}}],$ where $\mathbbm{1}_{\{\cdot\}}$ is the indicator function and $t$ is the time step.
From the definitions of the state and state-action visitation counts $\StateOccup^\discount_\policy$ and $\StateActionOccup^\discount_\policy$, it is straightforward to deduce that $\StateActionOccup^\discount_\policy(\state,\action) = \policy_{\state,\action}\StateOccup^\discount_\policy(\state)$, where $\policy_{\state,\action} = \mathbb{P}[A_t=a | S_t = s] $. 
We use the visitation counts to prove in Section~$4$ that
\begin{align}
    H_\policy^\discount = \sum_{(\state,\action)\in \States \times \Actions}  -(\log \pi_{\state,\action}) \pi_{\state,\action} \StateOccup^\discount_\policy (\state) 
&= \sum_{(\state,\action)\in \States \times \Actions} -\log \frac{\StateActionOccup^\discount_\policy(\state,\action)}{\StateOccup^\discount_\policy(\state)} \StateActionOccup^\discount_\policy(\state,\action), \nonumber
\end{align}%
where the last inequality is obtained by using that $\pi_{\state,\action} = \StateActionOccup^\discount_\policy(\state,\action) /  \StateOccup^\discount_\policy(\state)$. We claim in the paper that $H^\discount_\policy$ is a \emph{concave} fucntion of the visitation counts. Thus, we want to show that the function $f(\StateActionOccup^\gamma_\policy, \StateOccup^\gamma_\policy) = \sum_{(\state,\action)\in \States \times \Actions} -\log \frac{\StateActionOccup^\discount_\policy(\state,\action)}{\StateOccup^\discount_\policy(\state)} \StateActionOccup^\discount_\policy(\state,\action)$ is concave. To this end, consider any $\lambda \in (0,1)$ and the two sets of variables $\StateActionOccup^\gamma_\policy, \StateOccup^\gamma_\policy$ and $\bar{\StateActionOccup}^\gamma_\policy, \bar{\StateOccup}^\gamma_\policy$. 
Then, we have the following result:
\begin{align}
    &f(\lambda \StateActionOccup^\discount_\policy + (1-\lambda)  \bar{\StateActionOccup}^\discount_\policy, \lambda \bar{\StateOccup}^\discount_\policy + (1-\lambda)  \bar{\StateOccup}^\discount_\policy) \nonumber\\
    &\quad= \sum_{(\state,\action)\in \States \times \Actions} -\log \frac{\lambda \StateActionOccup^\discount_\policy(\state,\action) + (1-\lambda) \bar{\StateActionOccup}^\discount_\policy(\state,\action)}{\lambda \StateOccup^\discount_\policy(\state) + (1-\lambda) \bar{\StateOccup}^\discount_\policy(\state,\action)} (\lambda \StateActionOccup^\discount_\policy(\state,\action) + (1-\lambda) \bar{\StateActionOccup}^\discount_\policy(\state,\action)) \nonumber \\
    &\quad \geq \sum_{(\state,\action)\in \States \times \Actions} -\lambda \StateActionOccup^\discount_\policy(\state,\action) \log \frac{\lambda \StateActionOccup^\discount_\policy(\state,\action)}{\lambda \StateOccup^\discount_\policy(\state,\action)} - (1-\lambda) \bar{\StateActionOccup}^\discount_\policy(\state,\action) \log \frac{(1-\lambda) \bar{\StateActionOccup}^\discount_\policy(\state,\action)}{(1-\lambda) \bar{\StateOccup}^\discount_\policy(\state,\action)} \nonumber\\
    &\quad = \sum_{(\state,\action)\in \States \times \Actions} -\lambda \StateActionOccup^\discount_\policy(\state,\action) \log \frac{ \StateActionOccup^\discount_\policy(\state,\action)}{ \StateOccup^\discount_\policy(\state,\action)} - (1-\lambda) \bar{\StateActionOccup}^\discount_\policy(\state,\action) \log \frac{ \bar{\StateActionOccup}^\discount_\policy(\state,\action)}{ \bar{\StateOccup}^\discount_\policy(\state,\action)} \nonumber\\
    &\quad = \lambda f(\StateActionOccup^\discount_\policy, \StateOccup^\discount_\policy) + (1-\lambda) f(\bar{\StateActionOccup}^\discount_\policy,\bar{\StateOccup}^\discount_\policy), \nonumber
\end{align}
where the first inequality is obtained by applying to the well-known \emph{log-sum inequality}, i.e., $$x_1 \log \frac{x_1}{y_1} + x_2 \log \frac{x_2}{y_2} \geq (x_1+x_2) \log \frac{x_1+x_2}{y_1+y_2},$$ for nonnegative numbers $x_1,x_2, y_1, y_2$. 
Specifically, we apply the substitution $x_1 =  \lambda \StateActionOccup^\discount_\policy$, $y_1 = \lambda \StateOccup^\discount_\policy$, $x_2 = (1-\lambda) \bar{\StateActionOccup}^\discount_\policy$, and $y_2 = (1-\lambda) \bar{\StateOccup}^\discount_\policy$.
Note that the inequality 
$$f(\lambda \StateActionOccup^\discount_\policy + (1-\lambda)  \bar{\StateActionOccup}^\discount_\policy, \lambda \bar{\StateOccup}^\discount_\policy + (1-\lambda)  \bar{\StateOccup}^\discount_\policy) \geq \lambda f(\StateActionOccup^\discount_\policy, \StateOccup^\discount_\policy) + (1-\lambda) f(\bar{\StateActionOccup}^\discount_\policy,\bar{\StateOccup}^\discount_\policy)$$
implies that $f(\StateActionOccup^\gamma_\policy, \StateOccup^\gamma_\policy)$ is concave in $\StateActionOccup^\gamma_\policy,$ and  $\StateOccup^\gamma_\policy$.

\paragraph{Bellman Flow Constraint.} For the visitation count variables to correspond to a valid policy generating actions in the POMDP $\Pomdp$ , $\StateActionOccup^\discount_\policy$ and $\StateOccup^\discount_\policy$ must satisfy the bellman flow constraint given by
\begin{align}
    \StateOccup^\discount_{\policy} (\state)  &=
    \mathbb{E}_{S^\policy_t} \Big[ \sum_{t=0}^\infty \discount^t \mathbbm{1}_{\{S^\policy_t = \state\} }\Big] \nonumber \\
    &= \Initdist(\state) + \discount \mathbb{E}_{S^\policy_t} \Big[ \sum_{t=0}^\infty \discount^{t} \mathbbm{1}_{\{S^\policy_{t+1} = \state\} }\Big] \nonumber \\
    &= \Initdist(\state) + \discount  \sum_{t=0}^\infty \sum_{\state' \in \States} \sum_{\action \in \Actions} \discount^t \Transition(\state | \state',\action) \mathbb{P}[S^\policy_t=\state', A^\policy_t = \action] \nonumber\\
    &= \Initdist(\state) + \discount \sum_{\state' \in \States} \sum_{\action \in \Actions} \Transition(\state | \state',\action) \StateActionOccup^\discount_{\policy} (\state',\action).\nonumber
\end{align}%
\paragraph{Nonlinear Reward Parameterization.} 
Consider the parameterization of the reward function as  $\Reward(\state,\action) \triangleq h_\Rewardvec (\state, \action)$, where $\Rewardvec \in \Real^d$ is an unknown parameter vector and $h$ is a nonlinear function of $\Rewardvec$. 
Then, at each iteration of Algorithm~$2$ in line~3, the gradient step can be computed $\nabla h_\Rewardvec (\Rewardvec^k)$ or the corresponding the subgradient in the case $h_\Rewardvec$ is not differentiable.
We recall that for a linear parameterization, the gradient is obtained by $\nabla f (\Rewardvec^{k}) = R^\phi_{\policy^k} - \hat{R}^\phi$. 
Furthermore, the reader can straightforwardly observe that $\texttt{SCPForward}$ remains practically unchanged as, given $\Rewardvec^k$, we only need to replace the linear term $\sum_{(\state,\action) \in \States \times \Actions} (\Rewardvec^k)^\mathrm{T} \phi(\state,\action)\StateActionOccup^\discount_\policy(\state,\action)$ in the cost function by $\sum_{(\state,\action) \in \States \times \Actions} (\Rewardvec^k)^{\mathrm{T}} h_{\Rewardvec^k}(\state,\action) \StateActionOccup^\discount_\policy(\state,\action)$, which is also linear in the state-action visitation count $\StateActionOccup^\discount_\policy$.
Thus, the only challenge in handling nonlinear parameterization is computing the corresponding gradient, which can be done by standard platforms such as TensorFlow and Torch. 
Note that nonlinear parameterization enables to learn more expressive reward functions that are typically represented by neural networks.

\section{Experimental Tasks} 
In this section, we first provide a detailed description of the POMDP models used in the benchmark. 
The simulations on the benchmark examples empirically demonstrate that side information alleviates the information asymmetry, and more memory leads to more performant policies. 
Then, we provide additional numerical simulations supporting the claim that $\texttt{SCPForward}$ is sound and yields better scalability than off-the-shelf solvers for the \emph{forward problem}, i.e., computing an optimal policy on a POMDP for a given reward function.

\subsection{Computation Resources and External Assets}
All the experiments of this paper were performed on a computer with an Intel Core $i9$-$9900$ CPU $3.1$GHz $\times 16$ processors and $31.2$ Gb of RAM. 
All the implementations are written and tested in Python $3.8$, and we attach the code with the supplementary material. 

\paragraph{Required Tools. } Our implementation requires the Python interface \emph{Stormpy} of \emph{Storm}~\cite{hensel2020probabilistic} and \emph{Gurobipy} of \emph{Gurobi} $9.1$~\cite{gurobi}. 
On one hand, we use \emph{Storm}, a tool for model checking, to parse POMDP file specifications, to compute the product POMDP with the finite state controller in order  to reduce the synthesis problem to the synthesis of memoryless policies, and to compute the set $\target$ of target states satisfying a specification $\varphi$ via graph preprocessing. 
On the other hand, we use \emph{Gurobi} to solve both the linearized problem in  ($7$) and the feasible solution of the Bellman flow constraint needed for the \emph{verification step}.

\paragraph{Off-The-Shelf Solvers for Forward Problem. } In order to show the scalability of the developed algorithm $\texttt{SCPForward}$, we compare it to state-of-the-art POMDP solvers $\texttt{SolvePOMDP}$~\cite{walraven2017accelerated}, $\texttt{SARSOP}$~\cite{kurniawati2008sarsop}, and  $\texttt{PRISM-POMDP}$~\cite{norman2017verification}. 
The solver $\texttt{SolvePOMDP}$ implements both exact and approximate value iterations via incremental pruning~\cite{Cassandra97incrementalpruning} combined with state-of-the-art vector pruning methods~\cite{walraven2017accelerated}. 
Finally, $\texttt{PrismPOMDP}$ discretizes the belief state and adopts a finite memory strategy to find an approximate solution of the forward problem. For all the solvers above, we use the default settings except from the timeout enforced to be $3600$ seconds. 
These solvers are not provided with our implementation. However, we provide the POMDP models that each of the solvers can straightforwardly use. Further details are provided in the readme files of our implementation.

\subsection{Benchmark Set}
We evaluate the proposed learning algorithm on several POMDP instances adapted from~\cite{junges2020enforcing}. 
We attached the modified instances in our code with the automatically generated models for each off-the-shelf solver that the reader can straightforwardly use to reproduce Table~\ref{tab:ToolComp}. 
The reader can refer to Table~\ref{tab:ToolComp} for the number of states, observations, and transitions of each environment of the benchmark set. 
In all the examples, we gather $10$ trajectories from an expert that can fully observe its current state in the environment and an expert having partial observation of the environment. 
Our algorithm learns reward functions from these trajectories under different memory policies and high-level side information.

\clearpage
\paragraph{Rocks Instance. } 
In the environment $\mathrm{Rocks}$, an agent navigates in a gridworld to sample at least one valuable rock (if a valuable rock is in the grid)  over the two possibly dangerous rocks, without any failures. 
When at least one valuable rock has been collected, or the agent realizes that all the rocks are dangerous, it needs to get to an exit point to terminate the mission. 
The partial observability is due to the agent can only observe if its current location is an exit point or a dangerous rock. Furthermore, the agent has noisy sensors enabling sampling neighbor cells. 

We consider three feature functions. 
The first feature provides a positive reward when reaching the exit point with at least one valuable rock or no rocks when all of them are dangerous. 
The second feature provides a negative reward when the agent is at the location of a dangerous rock. 
Finally, the third feature penalizes each action taken with a negative reward to promote reaching the exit point as soon as possible.

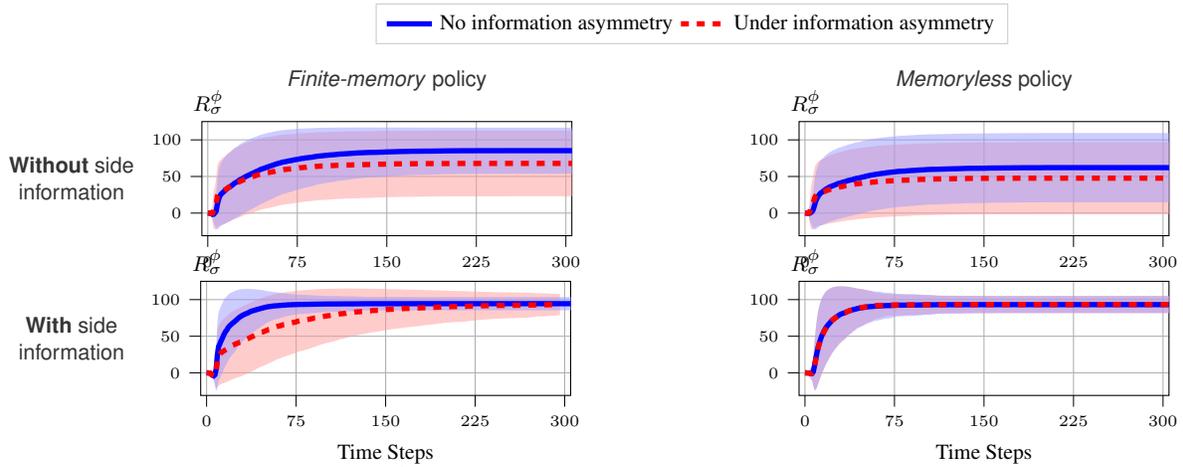
\begin{figure}[!hbt]
\centering
    \begin{subfigure}[t]{\textwidth}
        \centering\hspace*{2.4cm}\begin{tikzpicture}
\definecolor{color1}{HTML}{FF0000}
\definecolor{color0}{HTML}{0000FF}
\definecolor{color2}{HTML}{FF00FF}
\begin{customlegend}[legend columns=3,legend style={
  fill opacity=0.8,
  draw opacity=1,
  text opacity=1,
  at={(1.3,0.4)},
  anchor=south,
  draw=white!80!black,
  scale=0.40,
    font=\titlesize,
  mark options={scale=0.5},
},legend entries={No information asymmetry, Under information asymmetry}]

\addlegendimage{line width=2pt, color0}
\addlegendimage{line width=2pt, color1, dashed}
\end{customlegend}
\end{tikzpicture}
    \end{subfigure}\hfill%
    \vspace*{0.20cm}
    \centering
    \begin{subfigure}[t]{0.541\textwidth}
        \centering
        \centering\captionsetup{width=.95\linewidth}%
\begin{tikzpicture}

\definecolor{color1}{HTML}{FF0000}
\definecolor{color0}{HTML}{0000FF}

\begin{axis}[
legend cell align={left},
legend columns=2,
legend style={
  fill opacity=0.8,
  draw opacity=1,
  text opacity=1,
  at={(1.15,1.25)},
  anchor=south,
    ticklabel style = {font=\tiny},
  draw=white!80!black,
  font=\fontsize{7.4}{7.4}\selectfont,
  scale=0.40
},
title={$R_\policy^\phi$},
title style={yshift=-1.5ex,xshift=-15ex,font=\titlesize},
width=6.5cm,
height=3.1cm,
tick align=outside,
tick pos=left,
no markers,
every axis plot/.append style={line width=2pt},
x grid style={white!69.0196078431373!black},
xlabel={\footnotesize{Time steps}},
xlabel={\textcolor{gray!40!black}{\arial \textit{Finite-memory} policy}},
xlabel style={font=\titlesize,yshift=18ex},
xmajorgrids,
xmin=-4.95, xmax=305,
xtick style={color=black},
y grid style={white!69.0196078431373!black},
ylabel={\footnotesize{Mean  accumulated  reward}},
ylabel={\textcolor{gray!40!black}{\arial\textbf{Without} side} \\\textcolor{gray!40!black}{\arial information}},
ylabel style={rotate=-90,align=center,font=\titlesize},
xtick={0,75,150,225,300},
ymajorgrids,
ymin=-30.8979836098955, ymax=125,
ytick style={color=black}
]

\addplot[color0] table[x=x,y=y,mark=none] {rock_sim_files/rock_mem10_trajsize10pomdp_irl.tex};
\addplot[color1,dashed] table[x=x,y=y,mark=none] {rock_sim_files/rock_mem10_trajsize10mdp_irl.tex};

\addplot [name path=upper1,draw=none] table[x=x,y expr=\thisrow{y}+\thisrow{err}]{rock_sim_files/rock_mem10_trajsize10mdp_irl.tex};
\addplot [name path=lower1,draw=none] table[x=x,y expr=\thisrow{y}-\thisrow{err}] {rock_sim_files/rock_mem10_trajsize10mdp_irl.tex};
\addplot [fill=color1!40,opacity=0.5] fill between[of=upper1 and lower1];

\addplot [name path=upper,draw=none] table[x=x,y expr=\thisrow{y}+\thisrow{err}] {rock_sim_files/rock_mem10_trajsize10pomdp_irl.tex};
\addplot [name path=lower,draw=none] table[x=x,y expr=\thisrow{y}-\thisrow{err}] {rock_sim_files/rock_mem10_trajsize10pomdp_irl.tex};
\addplot [fill=color0!40,opacity=0.5] fill between[of=upper and lower];

\end{axis}

\end{tikzpicture}
    \end{subfigure}%
    \begin{subfigure}[t]{0.459\textwidth}
        \centering
        \centering\captionsetup{width=.95\linewidth}%
\begin{tikzpicture}

\definecolor{color1}{HTML}{FF0000}
\definecolor{color0}{HTML}{0000FF}

\begin{axis}[
legend cell align={left},
legend columns=2,
legend style={
  fill opacity=0.8,
  draw opacity=1,
  text opacity=1,
  at={(0.5,1)},
  anchor=south,
  ticklabel style = {font=\tiny},
  draw=white!80!black,
  font=\fontsize{6}{6}\selectfont,
  scale=0.40
},
title={$R_\policy^\phi$},
title style={yshift=-1.5ex,xshift=-15ex,font=\titlesize},
width=6.5cm,
height=3.1cm,
tick align=outside,
tick pos=left,
no markers,
every axis plot/.append style={line width=2pt},
x grid style={white!69.0196078431373!black},
xlabel={\footnotesize{Time steps}},
xlabel={},
xlabel={\textcolor{gray!40!black}{\arial \textit{Memoryless} policy}},
xlabel style={font=\titlesize,yshift=18ex},
xtick={0,75,150,225,300},
xmajorgrids,
xmin=-4.95, xmax=305,
xtick style={color=black},
y grid style={white!69.0196078431373!black},
ylabel={\footnotesize{Mean  accumulated  reward}},
ylabel={},
ymajorgrids,
ymin=-30.8979836098955, ymax=125,
ytick style={color=black}
]

\addplot[color0] table[x=x,y=y,mark=none] {rock_sim_files/rock_mem1_trajsize10pomdp_irl.tex};
\addplot[color1,dashed] table[x=x,y=y,mark=none] {rock_sim_files/rock_mem1_trajsize10mdp_irl.tex};

\addplot [name path=upper1,draw=none] table[x=x,y expr=\thisrow{y}+\thisrow{err}]{rock_sim_files/rock_mem1_trajsize10mdp_irl.tex};
\addplot [name path=lower1,draw=none] table[x=x,y expr=\thisrow{y}-\thisrow{err}] {rock_sim_files/rock_mem1_trajsize10mdp_irl.tex};
\addplot [fill=color1!40,opacity=0.5] fill between[of=upper1 and lower1];

\addplot [name path=upper,draw=none] table[x=x,y expr=\thisrow{y}+\thisrow{err}] {rock_sim_files/rock_mem1_trajsize10pomdp_irl.tex};
\addplot [name path=lower,draw=none] table[x=x,y expr=\thisrow{y}-\thisrow{err}] {rock_sim_files/rock_mem1_trajsize10pomdp_irl.tex};
\addplot [fill=color0!40,opacity=0.5] fill between[of=upper and lower];

\end{axis}

\end{tikzpicture}
    \end{subfigure}%
    \vfill
    \vspace{-0.50cm}
    \centering
        \begin{subfigure}[t]{0.541\textwidth}
        \centering
        \centering\captionsetup{width=.95\linewidth}%
\begin{tikzpicture}

\definecolor{color1}{HTML}{FF0000}
\definecolor{color0}{HTML}{0000FF}

\begin{axis}[
legend cell align={left},
legend columns=2,
legend style={
  fill opacity=0.8,
  draw opacity=1,
  text opacity=1,
  at={(0.5,1)},
  anchor=south,
  ticklabel style = {font=\tiny},
  draw=white!80!black,
  font=\fontsize{6}{6}\selectfont,
  scale=0.40
},
title={\textbf{Finite-memory} policy \emph{with} side information},
title={$R_\policy^\phi$},
title style={yshift=-1.5ex,xshift=-15ex,font=\titlesize},
width=6.5cm,
height=3.1cm,
tick align=outside,
tick pos=left,
no markers,
every axis plot/.append style={line width=2pt},
x grid style={white!69.0196078431373!black},
xlabel={\titlesize{Time Steps}},
xmajorgrids,
xmin=-4.95, xmax=305,
xtick style={color=black},
y grid style={white!69.0196078431373!black},
xtick={0,75,150,225,300},
ylabel={\footnotesize{Mean accumulated  reward}},
ylabel={\titlesize{$R_\policy^\phi$}},
ylabel={\textcolor{gray!40!black}{\;\;\,\arial\textbf{With} side\;\;\,} \\\textcolor{gray!40!black}{\arial information}},
ylabel style={rotate=-90,align=center,font=\titlesize},
ymajorgrids,
ymin=-30.8979836098955, ymax=125,
ytick style={color=black}
]

\addplot[color0] table[x=x,y=y,mark=none] {rock_sim_files/rock_mem10_trajsize10pomdp_irl_si.tex};
\addplot[color1,dashed] table[x=x,y=y,mark=none] {rock_sim_files/rock_mem10_trajsize10mdp_irl_si.tex};

\addplot [name path=upper1,draw=none] table[x=x,y expr=\thisrow{y}+\thisrow{err}]{rock_sim_files/rock_mem10_trajsize10mdp_irl_si.tex};
\addplot [name path=lower1,draw=none] table[x=x,y expr=\thisrow{y}-\thisrow{err}] {rock_sim_files/rock_mem10_trajsize10mdp_irl_si.tex};
\addplot [fill=color1!40,opacity=0.5] fill between[of=upper1 and lower1];

\addplot [name path=upper,draw=none] table[x=x,y expr=\thisrow{y}+\thisrow{err}] {rock_sim_files/rock_mem10_trajsize10pomdp_irl_si.tex};
\addplot [name path=lower,draw=none] table[x=x,y expr=\thisrow{y}-\thisrow{err}] {rock_sim_files/rock_mem10_trajsize10pomdp_irl_si.tex};
\addplot [fill=color0!40,opacity=0.5] fill between[of=upper and lower];

\end{axis}

\end{tikzpicture}
    \end{subfigure}%
    \begin{subfigure}[t]{0.459\textwidth}
        \centering
        \centering\captionsetup{width=.95\linewidth}%
\begin{tikzpicture}

\definecolor{color1}{HTML}{FF0000}
\definecolor{color0}{HTML}{0000FF}
\begin{axis}[
legend cell align={left},
legend columns=2,
legend style={
  fill opacity=0.8,
  draw opacity=1,
  text opacity=1,
  at={(0.5,1)},
  anchor=south,
  ticklabel style = {font=\tiny},
  draw=white!80!black,
  font=\fontsize{6}{6}\selectfont,
  scale=0.40
},
title={$R_\policy^\phi$},
title style={yshift=-1.5ex,xshift=-15ex,font=\titlesize},
width=6.5cm,
height=3.1cm,
tick align=outside,
tick pos=left,
no markers,
every axis plot/.append style={line width=2pt},
x grid style={white!69.0196078431373!black},
xlabel={\titlesize{Time Steps}},
xmajorgrids,
xmin=-4.95, xmax=305,
xtick style={color=black},
y grid style={white!69.0196078431373!black},
ylabel={\footnotesize{Mean  accumulated  reward}},
ylabel={},
ymajorgrids,
xtick={0,75,150,225,300},
ymin=-30.8979836098955, ymax=125,
ytick style={color=black}
]

\addplot[color0] table[x=x,y=y,mark=none] {rock_sim_files/rock_mem1_trajsize10pomdp_irl_si.tex};
\addplot[color1,dashed] table[x=x,y=y,mark=none] {rock_sim_files/rock_mem1_trajsize10mdp_irl_si.tex};

\addplot [name path=upper1,draw=none] table[x=x,y expr=\thisrow{y}+\thisrow{err}]{rock_sim_files/rock_mem1_trajsize10mdp_irl_si.tex};
\addplot [name path=lower1,draw=none] table[x=x,y expr=\thisrow{y}-\thisrow{err}] {rock_sim_files/rock_mem1_trajsize10mdp_irl_si.tex};
\addplot [fill=color1!40,opacity=0.5] fill between[of=upper1 and lower1];

\addplot [name path=upper,draw=none] table[x=x,y expr=\thisrow{y}+\thisrow{err}] {rock_sim_files/rock_mem1_trajsize10pomdp_irl_si.tex};
\addplot [name path=lower,draw=none] table[x=x,y expr=\thisrow{y}-\thisrow{err}] {rock_sim_files/rock_mem1_trajsize10pomdp_irl_si.tex};
\addplot [fill=color0!40,opacity=0.5] fill between[of=upper and lower];
\end{axis}

\end{tikzpicture}
    \end{subfigure}%
    \vspace*{-0.47cm}
    \caption{Representative results on the $\mathrm{Rock}$ example showing the reward of the policies under the true reward function ($R_\policy^\phi$) versus the time steps.}
    \label{fig:rock}
\end{figure}%

We compare scenarios with no side information and the a priori knowledge on the task such as \emph{the agent eventually reaches an exit point with a probability greater than $0.995$.} Figure~\ref{fig:rock} supports our claim that side information indeed alleviates the information asymmetry between the expert and the agent.
Additionally, we also observe that incorporating memory leads to more performant policies in terms of the mean accumulated reward.

\clearpage

\paragraph{Obstacle Instance. } 
In the environment $\mathrm{Obstacle}[n]$, an agent must find an exit in a gridworld without colliding with any of the five static obstacles in the grid. 
The agent only observes whether the current position is an obstacle or an exit state. The parameter $n$ specifies the dimension of the grid.

Similar to the $\mathrm{Rocks}$ example, the agent receives a positive reward if it successfully exits the gridworld and a negative reward for every taken action or colliding with an obstacle.

As for the side information, we specify in temporal logic that while learning the reward, \emph{the agent should not collide any obstacles until it reaches an exit point with a probability greater than $0.9$.}

\begin{figure}[!hbt]
\centering
    \begin{subfigure}[t]{\textwidth}
        \centering\hspace*{2.4cm}\begin{tikzpicture}
\definecolor{color1}{HTML}{FF0000}
\definecolor{color0}{HTML}{0000FF}
\definecolor{color2}{HTML}{FF00FF}
\begin{customlegend}[legend columns=3,legend style={
  fill opacity=0.8,
  draw opacity=1,
  text opacity=1,
  at={(1.3,0.4)},
  anchor=south,
  draw=white!80!black,
  scale=0.40,
    font=\titlesize,
  mark options={scale=0.5},
},legend entries={No information asymmetry, Under information asymmetry}]

\addlegendimage{line width=2pt, color0}
\addlegendimage{line width=2pt, color1, dashed}
\end{customlegend}
\end{tikzpicture}
    \end{subfigure}\hfill%
    \vspace*{0.20cm}
    \centering
    \begin{subfigure}[t]{0.541\textwidth}
        \centering
        \centering\captionsetup{width=.95\linewidth}%
\begin{tikzpicture}

\definecolor{color1}{HTML}{FF0000}
\definecolor{color0}{HTML}{0000FF}

\begin{axis}[
legend cell align={left},
legend columns=2,
legend style={
  fill opacity=0.8,
  draw opacity=1,
  text opacity=1,
  at={(1.15,1.25)},
  anchor=south,
    ticklabel style = {font=\tiny},
  draw=white!80!black,
  font=\fontsize{7.4}{7.4}\selectfont,
  scale=0.40
},
title={$R_\policy^\phi$},
title style={yshift=-1.5ex,xshift=-15ex,font=\titlesize},
width=6.5cm,
height=3.1cm,
tick align=outside,
tick pos=left,
no markers,
every axis plot/.append style={line width=2pt},
x grid style={white!69.0196078431373!black},
xlabel={\footnotesize{Time steps}},
xlabel={\textcolor{gray!40!black}{\arial \textit{Finite-memory} policy}},
xlabel style={font=\titlesize,yshift=18ex},
xmajorgrids,
xmin=-4.95, xmax=105,
xtick style={color=black},
y grid style={white!69.0196078431373!black},
ylabel={\footnotesize{Mean  accumulated  reward}},
ylabel={\textcolor{gray!40!black}{\arial\textbf{Without} side} \\\textcolor{gray!40!black}{\arial information}},
ylabel style={rotate=-90,align=center,font=\titlesize},
xtick={0,25,50,75,100},
ymajorgrids,
ymin=-200.8979836098955, ymax=550,
ytick style={color=black}
]

\addplot[color0] table[x=x,y=y,mark=none] {obstacle_sim_files/obstacle_mem5_trajsize10pomdp_irl.tex};
\addplot[color1,dashed] table[x=x,y=y,mark=none] {obstacle_sim_files/obstacle_mem5_trajsize10mdp_irl.tex};

\addplot [name path=upper1,draw=none] table[x=x,y expr=\thisrow{y}+\thisrow{err}]{obstacle_sim_files/obstacle_mem5_trajsize10mdp_irl.tex};
\addplot [name path=lower1,draw=none] table[x=x,y expr=\thisrow{y}-\thisrow{err}] {obstacle_sim_files/obstacle_mem5_trajsize10mdp_irl.tex};
\addplot [fill=color1!40,opacity=0.5] fill between[of=upper1 and lower1];

\addplot [name path=upper,draw=none] table[x=x,y expr=\thisrow{y}+\thisrow{err}] {obstacle_sim_files/obstacle_mem5_trajsize10pomdp_irl.tex};
\addplot [name path=lower,draw=none] table[x=x,y expr=\thisrow{y}-\thisrow{err}] {obstacle_sim_files/obstacle_mem5_trajsize10pomdp_irl.tex};
\addplot [fill=color0!40,opacity=0.5] fill between[of=upper and lower];

\end{axis}

\end{tikzpicture}
    \end{subfigure}%
    \begin{subfigure}[t]{0.459\textwidth}
        \centering
        \centering\captionsetup{width=.95\linewidth}%
\begin{tikzpicture}

\definecolor{color1}{HTML}{FF0000}
\definecolor{color0}{HTML}{0000FF}

\begin{axis}[
legend cell align={left},
legend columns=2,
legend style={
  fill opacity=0.8,
  draw opacity=1,
  text opacity=1,
  at={(0.5,1)},
  anchor=south,
  ticklabel style = {font=\tiny},
  draw=white!80!black,
  font=\fontsize{6}{6}\selectfont,
  scale=0.40
},
title={$R_\policy^\phi$},
title style={yshift=-1.5ex,xshift=-15ex,font=\titlesize},
width=6.5cm,
height=3.1cm,
tick align=outside,
tick pos=left,
no markers,
every axis plot/.append style={line width=2pt},
x grid style={white!69.0196078431373!black},
xlabel={\footnotesize{Time steps}},
xlabel={},
xlabel={\textcolor{gray!40!black}{\arial \textit{Memoryless} policy}},
xlabel style={font=\titlesize,yshift=18ex},
xtick={0,25,50,75,100},
xmajorgrids,
xmin=-4.95, xmax=105,
xtick style={color=black},
y grid style={white!69.0196078431373!black},
ylabel={\footnotesize{Mean  accumulated  reward}},
ylabel={},
ymajorgrids,
ymin=-200.8979836098955, ymax=550,
ytick style={color=black}
]

\addplot[color0] table[x=x,y=y,mark=none] {obstacle_sim_files/obstacle_mem1_trajsize10pomdp_irl.tex};
\addplot[color1,dashed] table[x=x,y=y,mark=none] {obstacle_sim_files/obstacle_mem1_trajsize10mdp_irl.tex};

\addplot [name path=upper1,draw=none] table[x=x,y expr=\thisrow{y}+\thisrow{err}]{obstacle_sim_files/obstacle_mem1_trajsize10mdp_irl.tex};
\addplot [name path=lower1,draw=none] table[x=x,y expr=\thisrow{y}-\thisrow{err}] {obstacle_sim_files/obstacle_mem1_trajsize10mdp_irl.tex};
\addplot [fill=color1!40,opacity=0.5] fill between[of=upper1 and lower1];

\addplot [name path=upper,draw=none] table[x=x,y expr=\thisrow{y}+\thisrow{err}] {obstacle_sim_files/obstacle_mem1_trajsize10pomdp_irl.tex};
\addplot [name path=lower,draw=none] table[x=x,y expr=\thisrow{y}-\thisrow{err}] {obstacle_sim_files/obstacle_mem1_trajsize10pomdp_irl.tex};
\addplot [fill=color0!40,opacity=0.5] fill between[of=upper and lower];

\end{axis}

\end{tikzpicture}
    \end{subfigure}%
    \vfill
    \vspace{-0.50cm}
    \centering
        \begin{subfigure}[t]{0.541\textwidth}
        \centering
        \centering\captionsetup{width=.95\linewidth}%
\begin{tikzpicture}

\definecolor{color1}{HTML}{FF0000}
\definecolor{color0}{HTML}{0000FF}

\begin{axis}[
legend cell align={left},
legend columns=2,
legend style={
  fill opacity=0.8,
  draw opacity=1,
  text opacity=1,
  at={(0.5,1)},
  anchor=south,
  ticklabel style = {font=\tiny},
  draw=white!80!black,
  font=\fontsize{6}{6}\selectfont,
  scale=0.40
},
title={\textbf{Finite-memory} policy \emph{with} side information},
title={$R_\policy^\phi$},
title style={yshift=-1.5ex,xshift=-15ex,font=\titlesize},
width=6.5cm,
height=3.1cm,
tick align=outside,
tick pos=left,
no markers,
every axis plot/.append style={line width=2pt},
x grid style={white!69.0196078431373!black},
xlabel={\titlesize{Time Steps}},
xmajorgrids,
xmin=-4.95, xmax=105,
xtick style={color=black},
y grid style={white!69.0196078431373!black},
xtick={0,25,50,75,100},
ylabel={\footnotesize{Mean accumulated  reward}},
ylabel={\titlesize{$R_\policy^\phi$}},
ylabel={\textcolor{gray!40!black}{\;\;\,\arial\textbf{With} side\;\;\,} \\\textcolor{gray!40!black}{\arial information}},
ylabel style={rotate=-90,align=center,font=\titlesize},
ymajorgrids,
ymin=-200.8979836098955, ymax=550,
ytick style={color=black}
]

\addplot[color0] table[x=x,y=y,mark=none] {obstacle_sim_files/obstacle_mem5_trajsize10pomdp_irl_si.tex};
\addplot[color1,dashed] table[x=x,y=y,mark=none] {obstacle_sim_files/obstacle_mem5_trajsize10mdp_irl_si.tex};

\addplot [name path=upper1,draw=none] table[x=x,y expr=\thisrow{y}+\thisrow{err}]{obstacle_sim_files/obstacle_mem5_trajsize10mdp_irl_si.tex};
\addplot [name path=lower1,draw=none] table[x=x,y expr=\thisrow{y}-\thisrow{err}] {obstacle_sim_files/obstacle_mem5_trajsize10mdp_irl_si.tex};
\addplot [fill=color1!40,opacity=0.5] fill between[of=upper1 and lower1];

\addplot [name path=upper,draw=none] table[x=x,y expr=\thisrow{y}+\thisrow{err}] {obstacle_sim_files/obstacle_mem5_trajsize10pomdp_irl_si.tex};
\addplot [name path=lower,draw=none] table[x=x,y expr=\thisrow{y}-\thisrow{err}] {obstacle_sim_files/obstacle_mem5_trajsize10pomdp_irl_si.tex};
\addplot [fill=color0!40,opacity=0.5] fill between[of=upper and lower];

\end{axis}

\end{tikzpicture}
    \end{subfigure}%
    \begin{subfigure}[t]{0.459\textwidth}
        \centering
        \centering\captionsetup{width=.95\linewidth}%
\begin{tikzpicture}

\definecolor{color1}{HTML}{FF0000}
\definecolor{color0}{HTML}{0000FF}
\begin{axis}[
legend cell align={left},
legend columns=2,
legend style={
  fill opacity=0.8,
  draw opacity=1,
  text opacity=1,
  at={(0.5,1)},
  anchor=south,
  ticklabel style = {font=\tiny},
  draw=white!80!black,
  font=\fontsize{6}{6}\selectfont,
  scale=0.40
},
title={$R_\policy^\phi$},
title style={yshift=-1.5ex,xshift=-15ex,font=\titlesize},
width=6.5cm,
height=3.1cm,
tick align=outside,
tick pos=left,
no markers,
every axis plot/.append style={line width=2pt},
x grid style={white!69.0196078431373!black},
xlabel={\titlesize{Time Steps}},
xmajorgrids,
xmin=-4.95, xmax=105,
xtick style={color=black},
y grid style={white!69.0196078431373!black},
ylabel={\footnotesize{Mean  accumulated  reward}},
ylabel={},
ymajorgrids,
xtick={0,25,50,75,100},
ymin=-200.8979836098955, ymax=550,
ytick style={color=black}
]

\addplot[color0] table[x=x,y=y,mark=none] {obstacle_sim_files/obstacle_mem1_trajsize10pomdp_irl_si.tex};
\addplot[color1,dashed] table[x=x,y=y,mark=none] {obstacle_sim_files/obstacle_mem1_trajsize10mdp_irl_si.tex};

\addplot [name path=upper1,draw=none] table[x=x,y expr=\thisrow{y}+\thisrow{err}]{obstacle_sim_files/obstacle_mem1_trajsize10mdp_irl_si.tex};
\addplot [name path=lower1,draw=none] table[x=x,y expr=\thisrow{y}-\thisrow{err}] {obstacle_sim_files/obstacle_mem1_trajsize10mdp_irl_si.tex};
\addplot [fill=color1!40,opacity=0.5] fill between[of=upper1 and lower1];

\addplot [name path=upper,draw=none] table[x=x,y expr=\thisrow{y}+\thisrow{err}] {obstacle_sim_files/obstacle_mem1_trajsize10pomdp_irl_si.tex};
\addplot [name path=lower,draw=none] table[x=x,y expr=\thisrow{y}-\thisrow{err}] {obstacle_sim_files/obstacle_mem1_trajsize10pomdp_irl_si.tex};
\addplot [fill=color0!40,opacity=0.5] fill between[of=upper and lower];
\end{axis}

\end{tikzpicture}
    \end{subfigure}%
    \vspace*{-0.47cm}
    \caption{Representative results on the $\mathrm{Obstacle}$ example showing the reward of the policies under the true reward function ($R_\policy^\phi$) versus the time steps.}
    \label{fig:obstacle}
\end{figure}
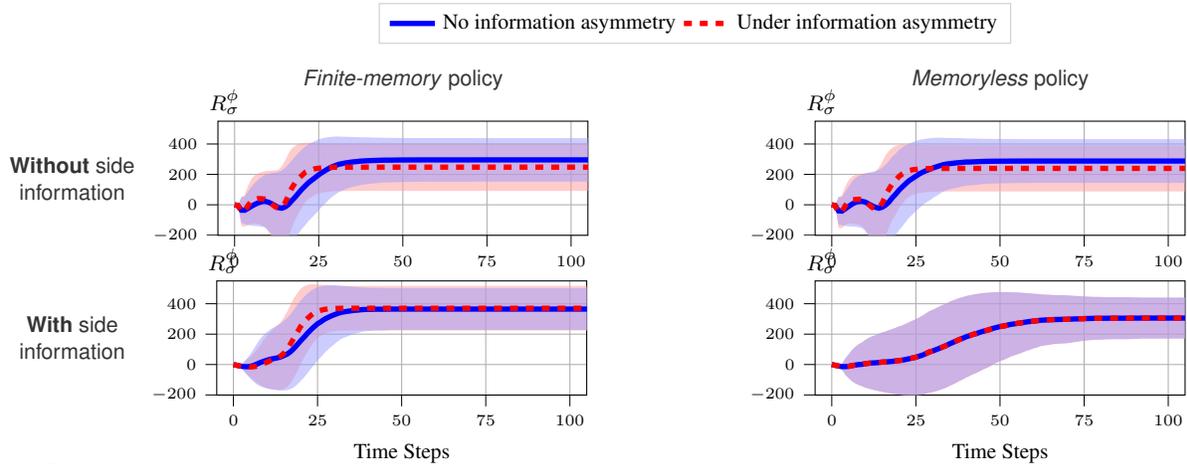%

Similar to the $\mathrm{Maze}$ and $\mathrm{Rock}$ examples,
Figure~\ref{fig:obstacle} supports our claim that side information alleviates the information asymmetry and memory leads to more performant policies.

\clearpage

\paragraph{Evade Instance. } 
$\mathrm{Evade}[n,r,slip]$ is a turn-based game where the agent must reach a destination without being intercepted by a faster player. 
The player cannot access the top row of the grid. 
Further, the agent can only observe the player if it is within a fixed radius from its current location and upon calling the action \emph{scan}. The parameters $n$, $r$, and $slip$ specify the dimension of the grid, the view radius, and the slippery probability, respectively.

The feature functions are defined such that the agent receives a positive reward if at the destination, a high negative reward if it is intercept by the player, and a small negative reward for each action taken, including the \emph{scan} action.

\begin{figure}[!hbt]
\centering
    \begin{subfigure}[t]{\textwidth}
        \centering\hspace*{1.4cm}\begin{tikzpicture}
\definecolor{color1}{HTML}{FF0000}
\definecolor{color0}{HTML}{0000FF}
\begin{customlegend}[legend columns=2,legend style={
  fill opacity=0.8,
  draw opacity=1,
  text opacity=1,
  at={(0.0,110.2)},
  anchor=south,
  draw=white!80!black,
  scale=0.40,
  mark options={scale=0.5},
},legend entries={With side information, Without side information}]
\addlegendimage{line width=2pt, color0}
\addlegendimage{line width=2pt, color1,dashed}
\end{customlegend}
\end{tikzpicture}
    \end{subfigure}\hfill%
    \vspace*{0.20cm}
    \centering
    \begin{subfigure}[t]{\textwidth}
        \centering
        \centering\captionsetup{width=.95\linewidth}%
\begin{tikzpicture}

\definecolor{color0}{HTML}{FF0000}
\definecolor{color1}{HTML}{0000FF}
\begin{axis}[
legend cell align={left},
legend columns=2,
legend style={
  fill opacity=0.8,
  draw opacity=1,
  text opacity=1,
  at={(0.5,1)},
  anchor=south,
  draw=white!80!black,
  font=\fontsize{6}{6}\selectfont,
  scale=0.40
},
title={$R_\policy^\phi$},
title={},
title style={yshift=-1.5ex,xshift=-15ex,font=\titlesize},
width=\textwidth,
height=0.4\textwidth,
tick align=outside,
tick pos=left,
no markers,
every axis plot/.append style={line width=2pt},
x grid style={white!69.0196078431373!black},
xlabel={{Time Steps}},
xmajorgrids,
xmin=-4.95, xmax=110,
xtick style={color=black},
y grid style={white!69.0196078431373!black},
ylabel={{Mean  accumulated  reward}},
ymajorgrids,
xtick={0,25,50,75,100},
ymin=-10, ymax=25,
ticklabel style = {font=\normalsize}
]

\addplot[color0,dashed] table[x=x,y=y,mark=none] {evade_sim_files/evade_mem1_trajsize10mdp_irl.tex};
\addplot[color1] table[x=x,y=y,mark=none] {evade_sim_files/evade_mem1_trajsize10mdp_irl_si.tex};

\addplot [name path=upper1,draw=none] table[x=x,y expr=\thisrow{y}+\thisrow{err}]{evade_sim_files/evade_mem1_trajsize10mdp_irl_si.tex};
\addplot [name path=lower1,draw=none] table[x=x,y expr=\thisrow{y}-\thisrow{err}] {evade_sim_files/evade_mem1_trajsize10mdp_irl_si.tex};
\addplot [fill=color1!40,opacity=0.5] fill between[of=upper1 and lower1];

\addplot [name path=upper,draw=none] table[x=x,y expr=\thisrow{y}+\thisrow{err}] {evade_sim_files/evade_mem1_trajsize10mdp_irl.tex};
\addplot [name path=lower,draw=none] table[x=x,y expr=\thisrow{y}-\thisrow{err}] {evade_sim_files/evade_mem1_trajsize10mdp_irl.tex};
\addplot [fill=color0!40,opacity=0.5] fill between[of=upper and lower];
\end{axis}

\end{tikzpicture}
    \end{subfigure}%
    \caption{Representative results on the $\mathrm{Evade}$ example showing the reward of the policies under the true reward function ($R_\policy^\phi$) versus the time steps.}
    \label{fig:evade}
\end{figure}
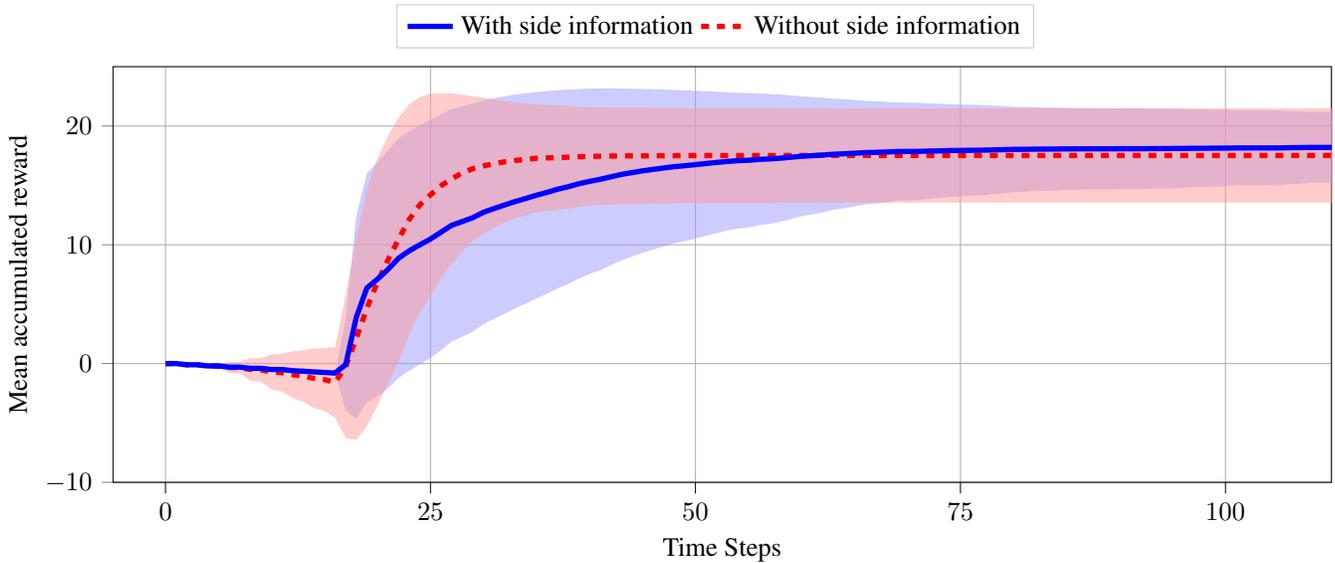

As for the side information, we specify in temporal logic that while learning the reward, \emph{the agent must reach an exit point with probability greater than $0.98$.} 

Figure~\ref{fig:evade} shows that learning with side information provides higher reward than without side information. Besides, there is less randomness in the policy with side information compared to the policy without side information. Specifically, the standard deviation of the policy with side information is significantly smaller than the policy without side information.  

We did not discuss the impact of different memory size policies in this example since the performance of the memoryless policy is already near-optimal, as the policy obtains the same reward as $\texttt{SARSOP}$ (see Table~\ref{tab:ToolComp} for a reference.
Specifically, we observe that the optimal policy on the underlying MDP yields comparable policies to the optimal memoryless policy on the POMDP. 
As a consequence, we observe that the information asymmetry between the expert and the agent does not hold here either, and the learned policies obtain a similar performance.

\clearpage

\paragraph{Intercept Instance. } 
$\mathrm{Intercept}[n,r,slip]$ is a variant of $\mathrm{Evade}$ where the agent must intercept another player who is trying to exit the gridworld. 
The agent can move in $8$ directions and can only observe the player if it is within a fixed radius from the agent's current position when the action \emph{scan} is performed. 
Besides, the agent has a camera that enables it to observe all cells from west to east from the center of the gridworld. 
In contrast, the player can only move in $4$ directions. The parameters $n$, $r$, and $slip$ specify the dimension of the grid, the view radius, and the slippery probability, respectively.

We consider three feature functions to parameterize the unknown reward. The first feature provides a positive reward to the agent upon intercepting the player. 
The second feature penalizes the agent if the player exits the gridworld. 
The third feature imposes a penalty cost for each action taken.

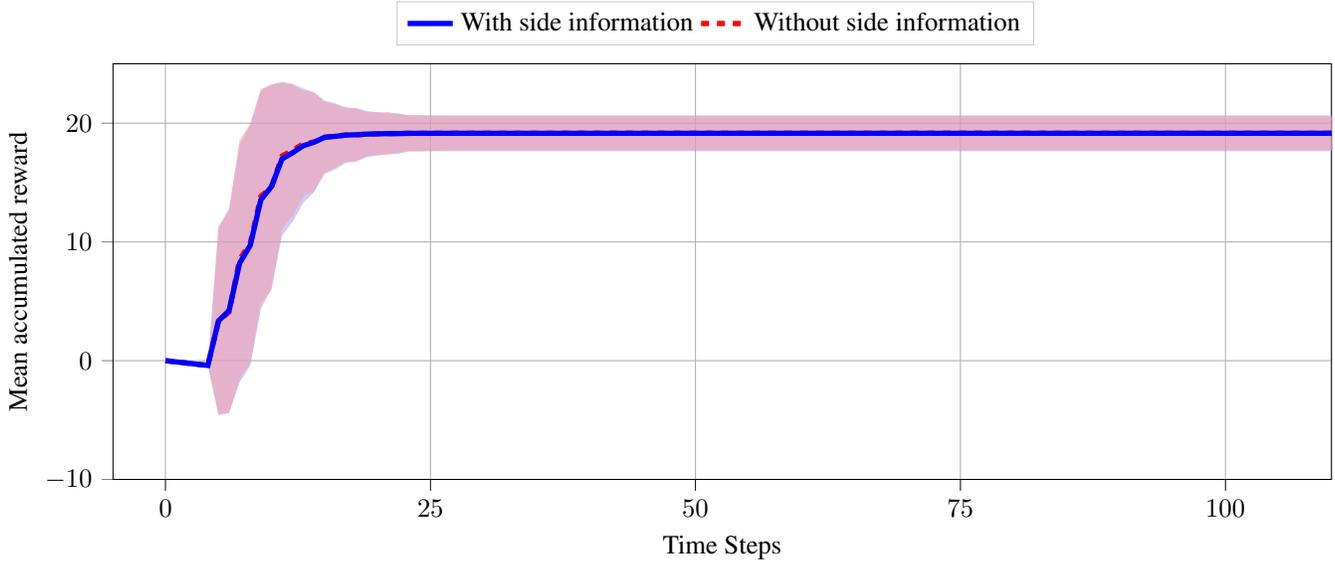
\begin{figure}[!hbt]
\centering
    \begin{subfigure}[t]{\textwidth}
        \centering\hspace*{1.4cm}\begin{tikzpicture}
\definecolor{color1}{HTML}{FF0000}
\definecolor{color0}{HTML}{0000FF}
\begin{customlegend}[legend columns=2,legend style={
  fill opacity=0.8,
  draw opacity=1,
  text opacity=1,
  at={(0.0,110.2)},
  anchor=south,
  draw=white!80!black,
  scale=0.40,
  mark options={scale=0.5},
},legend entries={With side information, Without side information}]
\addlegendimage{line width=2pt, color0}
\addlegendimage{line width=2pt, color1,dashed}
\end{customlegend}
\end{tikzpicture}
    \end{subfigure}\hfill%
    \vspace*{0.20cm}
    \centering
    \begin{subfigure}[t]{\textwidth}
        \centering
        \centering\captionsetup{width=.95\linewidth}%
\begin{tikzpicture}

\definecolor{color0}{HTML}{FF0000}
\definecolor{color1}{HTML}{0000FF}
\begin{axis}[
legend cell align={left},
legend columns=2,
legend style={
  fill opacity=0.8,
  draw opacity=1,
  text opacity=1,
  at={(0.5,1)},
  anchor=south,
  draw=white!80!black,
  font=\fontsize{6}{6}\selectfont,
  scale=0.40
},
title={$R_\policy^\phi$},
title={},
title style={yshift=-1.5ex,xshift=-15ex,font=\titlesize},
width=\textwidth,
height=0.4\textwidth,
tick align=outside,
tick pos=left,
no markers,
every axis plot/.append style={line width=2pt},
x grid style={white!69.0196078431373!black},
xlabel={{Time Steps}},
xmajorgrids,
xmin=-4.95, xmax=110,
xtick style={color=black},
y grid style={white!69.0196078431373!black},
ylabel={{Mean  accumulated  reward}},
ymajorgrids,
xtick={0,25,50,75,100},
ymin=-10, ymax=25,
ticklabel style = {font=\normalsize}
]

\addplot[color0,dashed] table[x=x,y=y,mark=none] {intercept_sim_files/intercept_mem1_trajsize10mdp_irl.tex};
\addplot[color1] table[x=x,y=y,mark=none] {intercept_sim_files/intercept_mem1_trajsize10mdp_irl_si.tex};

\addplot [name path=upper1,draw=none] table[x=x,y expr=\thisrow{y}+\thisrow{err}]{intercept_sim_files/intercept_mem1_trajsize10mdp_irl_si.tex};
\addplot [name path=lower1,draw=none] table[x=x,y expr=\thisrow{y}-\thisrow{err}] {intercept_sim_files/intercept_mem1_trajsize10mdp_irl_si.tex};
\addplot [fill=color1!40,opacity=0.5] fill between[of=upper1 and lower1];

\addplot [name path=upper,draw=none] table[x=x,y expr=\thisrow{y}+\thisrow{err}] {intercept_sim_files/intercept_mem1_trajsize10mdp_irl.tex};
\addplot [name path=lower,draw=none] table[x=x,y expr=\thisrow{y}-\thisrow{err}] {intercept_sim_files/intercept_mem1_trajsize10mdp_irl.tex};
\addplot [fill=color0!40,opacity=0.5] fill between[of=upper and lower];
\end{axis}

\end{tikzpicture}
    \end{subfigure}%
    \caption{Representative results on the $\mathrm{Intercept}$ example showing the reward of the policies under the true reward function ($R_\policy^\phi$) versus the time steps.}
    \label{fig:intercept}
\end{figure}

We encode the high-level side information as the temporal logic task specification \emph{Eventually intercept the player with probability greater than $0.98$}, i.e., the agent should eventually reach an observation where its location coincides with the player's location.

Figure~\ref{fig:intercept} demonstrates that side information does not improve the performance of the policy. 
This result is because memoryless policies are optimal in this example, and a combination of the given reward features can perfectly encode the temporal logic specifications, similar to the $\mathrm{Evade}$ example.

\clearpage

\paragraph{Avoid Instance. } 
$\mathrm{Avoid}[n,r,slip]$ is a variant of $\mathrm{Evade}$, where the agent must reach an exit point while avoiding being detected by two other moving players following certain predefined yet unknown routes. 
The agent can only observe the players if they are within a fixed radius from the agent's current position when the action \emph{scan} is performed.
Besides, with the players' speed being uncertain, their position in the routes can not be inferred by the agent. The parameters $n$, $r$, and $slip$ specify the dimension of the grid, the view radius, and the slippery probability, respectively.

We consider four feature functions to parameterize the unknown reward. The first feature provides a positive reward to the agent upon reaching the exit point. 
The second feature penalizes the agent if it collides with a player. 
The third feature penalizes the agent if it is detected by a player. 
The fourth feature imposes a penalty cost for each action taken.

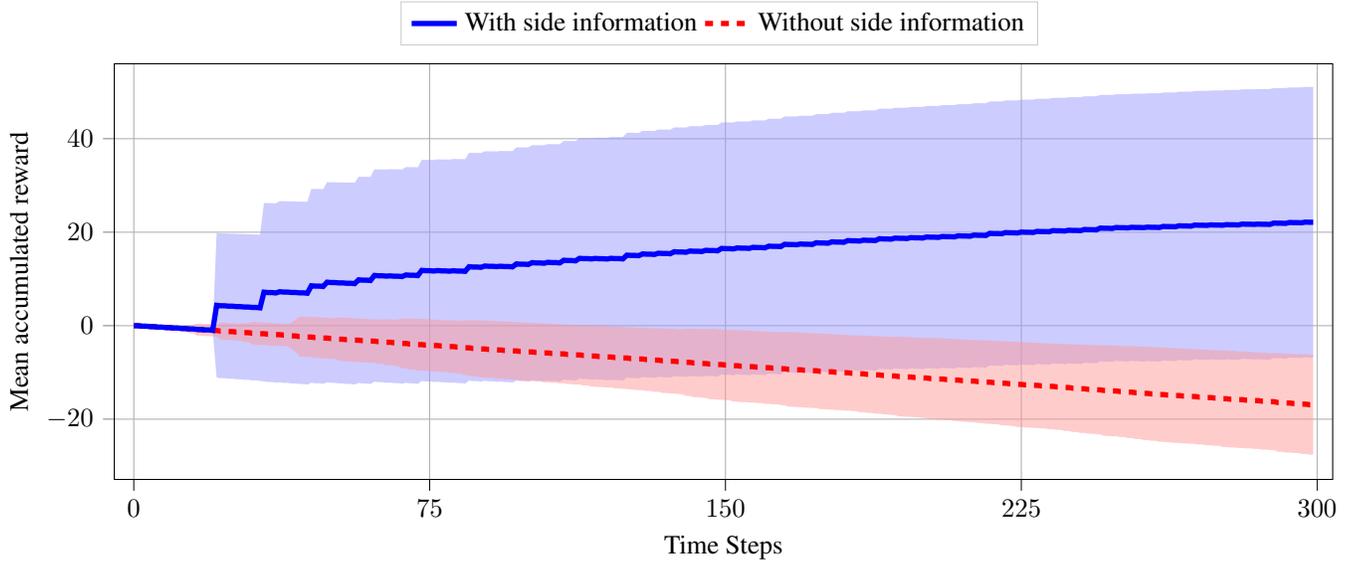
\begin{figure}[!hbt]
\centering
    \begin{subfigure}[t]{\textwidth}
        \centering\hspace*{1.4cm}\begin{tikzpicture}
\definecolor{color1}{HTML}{FF0000}
\definecolor{color0}{HTML}{0000FF}
\begin{customlegend}[legend columns=2,legend style={
  fill opacity=0.8,
  draw opacity=1,
  text opacity=1,
  at={(0.0,110.2)},
  anchor=south,
  draw=white!80!black,
  scale=0.40,
  mark options={scale=0.5},
},legend entries={With side information, Without side information}]
\addlegendimage{line width=2pt, color0}
\addlegendimage{line width=2pt, color1,dashed}
\end{customlegend}
\end{tikzpicture}
    \end{subfigure}\hfill%
    \vspace*{0.20cm}
    \centering
    \begin{subfigure}[t]{\textwidth}
        \centering
        \centering\captionsetup{width=.95\linewidth}%
\begin{tikzpicture}

\definecolor{color0}{HTML}{FF0000}
\definecolor{color1}{HTML}{0000FF}
\begin{axis}[
legend cell align={left},
legend columns=2,
legend style={
  fill opacity=0.8,
  draw opacity=1,
  text opacity=1,
  at={(0.5,1)},
  anchor=south,
  draw=white!80!black,
  font=\fontsize{6}{6}\selectfont,
  scale=0.40
},
title={$R_\policy^\phi$},
title={},
title style={yshift=-1.5ex,xshift=-15ex,font=\titlesize},
width=\textwidth,
height=0.4\textwidth,
tick align=outside,
tick pos=left,
no markers,
every axis plot/.append style={line width=2pt},
x grid style={white!69.0196078431373!black},
xlabel={{Time Steps}},
xmajorgrids,
xmin=-4.95, xmax=303.95,
xtick style={color=black},
y grid style={white!69.0196078431373!black},
ylabel={{Mean  accumulated  reward}},
ymajorgrids,
xtick={0,75,150,225,300},
ymin=-32.8979836098955, ymax=56.0315604206547,
ticklabel style = {font=\normalsize}
]

\addplot[color0,dashed] table[x=x,y=y,mark=none] {avoid_sim_files/avoid_mem1_trajsize10mdp_irl.tex};
\addplot[color1] table[x=x,y=y,mark=none] {avoid_sim_files/avoid_mem1_trajsize10mdp_irl_si.tex};

\addplot [name path=upper1,draw=none] table[x=x,y expr=\thisrow{y}+\thisrow{err}]{avoid_sim_files/avoid_mem1_trajsize10mdp_irl_si.tex};
\addplot [name path=lower1,draw=none] table[x=x,y expr=\thisrow{y}-\thisrow{err}] {avoid_sim_files/avoid_mem1_trajsize10mdp_irl_si.tex};
\addplot [fill=color1!40,opacity=0.5] fill between[of=upper1 and lower1];

\addplot [name path=upper,draw=none] table[x=x,y expr=\thisrow{y}+\thisrow{err}] {avoid_sim_files/avoid_mem1_trajsize10mdp_irl.tex};
\addplot [name path=lower,draw=none] table[x=x,y expr=\thisrow{y}-\thisrow{err}] {avoid_sim_files/avoid_mem1_trajsize10mdp_irl.tex};
\addplot [fill=color0!40,opacity=0.5] fill between[of=upper and lower];
\end{axis}

\end{tikzpicture}
    \end{subfigure}%
    \caption{Representative results on the $\mathrm{Avoid}$ example showing the reward of the policies under the true reward function ($R_\policy^\phi$) versus the time steps.}
    \label{fig:avoid}
\end{figure}

We encode the high-level side information as the temporal logic task specification \emph{avoid being detected until reaching the exit point with probability greater than $0.98$}.

Figure~\ref{fig:avoid} shows that the algorithm is unable to learn without side information while side information induces a learned policy that is  optimal. Specifically, the learned policy without side information seems to only focus on avoiding being detected and collision as the corresponding learned features were close to zero.
Thus, the agent using that policy does not move until sensing a player is closed by. As a consequence it gets always a negative reward from taking actions at each time step. 

\subsection{Summary of the Results}

\paragraph{Side Information Alleviates the Information Asymmetry. }
As mentioned in the submitted manuscript, side information can indeed alleviate the information asymmetry.
Specifically, we observe that if there is an information asymmetry in the \emph{forward problem}, i.e., the obtained reward from an optimal policy on the underlying POMDP is lower than from an optimal policy on the underlying fully observable MDP, incorporating side information in temporal logic specifications alleviates the information asymmetry between the expert and the agent.
For example, we can see the effects of such information asymmetry in the $\mathrm{Maze}$, $\mathrm{Rocks}$, $\mathrm{Obstacle}$, and $\mathrm{Avoid}$ examples.
In these examples, having partial observability reduces the obtained reward in the forward problem. 
The policies that do not incorporate side information into the learning procedure also obtain a lower reward under information asymmetry.

\paragraph{Memory Leads to More Performance Policies.}
Similarly to the side information, we also observe that if incorporating memory improves the performance of the learned policies, if it also improves the obtained reward in the forward problem, as seen in the $\mathrm{Maze}$, $\mathrm{Rocks}$, and $\mathrm{Obstacle}$ instances.
In Table~\ref{tab:ToolComp}, we can also see that incorporating memory helps to compute a better optimal policy in these examples, unlike computing a memoryless policy.

\clearpage

\subsection{\texttt{SCPForward} Yields Better Scalability.}

We support our claim that \texttt{SCPForward} yields better scalability through additional simulations on the benchmark examples by varying the size of the considered examples. Specifically, we demonstrate that on large-size problems, only $\texttt{SARSOP}$ attempts to solve the problems, and when it succeeds, its computation time is at least two orders of magnitude slower than $\texttt{SCPForward}$.

\begin{table*}[!hbt]
	\setlength{\tabcolsep}{4.5pt}
	\centering
	\scalebox{0.95}{%
		\begin{tabular}{@{}cccc|cc|cc|cc@{}}
			\toprule
			&  & & & \multicolumn{2}{c|}{$\texttt{SCPForward}$}   &\multicolumn{2}{c|}{$\texttt{SARSOP}$} & \multicolumn{2}{c}{$\texttt{SolvePOMDP}$}  \\
			Problem   & $|\States|$ & $|\States \times \Observations|$ & $|\Observations|$ & $R_\policy^\phi$ & Time (s) & $R_\policy^\phi$ & Time (s) & $R_\policy^\phi$ & Time (s) \\
			\midrule
			$\mathrm{Maze}$ & $17$  & $162$ & $11$ & $39.24$ &   $\mathbf{0.1}$ & $\mathbf{47.83}$ & $0.24$ &  $47.83$ & $0.33$\\
			$\mathrm{Maze}$ ($3$-$\mathrm{FSC})$ & $49$  & $777$ & $31$ & $44.98$ &   $\mathbf{0.6}$ & $\mathbf{47.83}$ & $0.24$ &  $\mathbf{47.83}$ & $0.33$\\
			$\mathrm{Maze}$ ($10$-$\mathrm{FSC})$ & $161$  & $2891$ & $101$ & $46.32$ &   $2.04$ & $\mathbf{47.83}$ & $\mathbf{0.24}$ &  $47.83$ & $0.33$\\
			$\mathrm{Obstacle}[10]$ & $102$  & $1126$ & $5$ & $19.71$ &   $8.79$ & $\mathbf{19.8}$ & $\mathbf{0.02}$ & $5.05$ & $3600$ \\
			$\mathrm{Obstacle}[10]$($5$-$\mathrm{FSC})$ & $679$  & $7545$ & $31$ & $19.77$ &   $38$ & $\mathbf{19.8}$ & $\mathbf{0.02}$ & $5.05$ & $3600$ \\
			$\mathrm{Obstacle}[25]$ & $627$  & $7306$ & $5$ & $19.59$ &   $14.22$ & $\textbf{19.8}$ & $\textbf{0.1}$ & $5.05$ & $3600$ \\
			$\mathrm{Rock}$ & $550$  & $4643$ & $67$ & $19.68$ &   $12.2$ & $\mathbf{19.83}$ & $\mathbf{0.05}$ &  $-$ & $-$\\
			$\mathrm{Rock}$ ($3$-$\mathrm{FSC})$ & $1648$  & $23203$ & $199$ & $19.8$ &   $15.25$ & $\mathbf{19.83}$ & $\mathbf{0.05}$ &  $-$ & $-$\\
			$\mathrm{Rock}$ ($5$-$\mathrm{FSC})$ & $2746$  & $41759$ & $331$ & $19.82$ &   $97.84$ & $\mathbf{19.83}$ & $\mathbf{0.05}$ &  $-$ & $-$\\
			$\mathrm{Intercept}[5,2,0]$ & $1321$  & $5021$ & $1025$ & $\mathbf{19.83}$ &   $\mathbf{10.28}$ & $\mathbf{19.83}$ & $13.71$ &  $-$ & $-$\\
			$\mathrm{Intercept}[5,2,0.1]$ & $1321$  & $7041$ & $1025$ & $\mathbf{19.81}$ &   $\mathbf{13.18}$ & $\mathbf{19.81}$ & $81.19$ &  $-$ & $-$\\
			$\mathrm{Evade}[5,2,0]$ & $2081$  & $13561$ & $1089$ & $\mathbf{97.3}$ &   $\mathbf{26.25}$ & $\mathbf{97.3}$ & $3600$ &  $-$ & $-$\\
			$\mathrm{Evade}[5,2,0.1]$ & $2081$  & $16761$ & $1089$ & $\mathbf{96.79}$ &   $\mathbf{26.25}$ & $95.28$ & $3600$ &  $-$ & $-$\\
			$\mathrm{Evade}[10,2,0]$ & $36361$  & $341121$ & $18383$ & $\mathbf{94.97}$ &   $\mathbf{3600}$ & $-$ & $-$ &  $-$ & $-$\\
			$\mathrm{Avoid}[4,2,0]$ & $2241$  & $5697$ & $1956$ & $\mathbf{9.86}$ &   $34.74$ & $\mathbf{9.86}$ & $\mathbf{9.19}$ &  $-$ & $-$\\
			$\mathrm{Avoid}[4,2,0.1]$ & $2241$  & $8833$ & $1956$ & $\mathbf{9.86}$ &   $\mathbf{14.63}$ & $\mathbf{9.86}$ & $210.47$ &  $-$ & $-$\\
			$\mathrm{Avoid}[7,2,0]$ & $19797$  & $62133$ & $3164$ & $\mathbf{9.72}$ &   $\mathbf{3503}$ & $-$ & $-$ &  $-$ & $-$\\
			\bottomrule
	\end{tabular}}
	\caption{Results for the benchmarks. On larger benchmarks (e.g., $\mathrm{Evade}$ and $\mathrm{Avoid}$), only $\texttt{SCPForward}$ can compute a locally optimal policy. We set the time-out to $3600$ seconds. An empty cell (denoted by $-$) represents the solver failed to compute any policy before the time-out.}
	\label{tab:ToolComp}
\end{table*}

We see that in Table~\ref{tab:ToolComp}, incorporating memory into policy on the $\mathrm{Maze}$, $\mathrm{Rock}$, and $\mathrm{Obstacle}$ benchmarks, allows $\texttt{SCPForward}$ to compute policies that are as optimal as $\texttt{SARSOP}$, which uses belief update rules. We recall that the comparisons with $\texttt{PrismPOMDP}$ were not reported in this table due to either having negative values in the reward function, which is not allowed by $\texttt{PrismPOMDP}$, or the solver runs out of memory before computing any policy.

\end{document}